\documentclass[journal]{IEEEtran}
\ifCLASSINFOpdf
\else
\fi
\usepackage{amsmath}
\usepackage{amsfonts}
\usepackage{subfigure}
\usepackage{amssymb}
\usepackage{float}
\usepackage[linesnumbered,ruled,vlined]{algorithm2e}
\usepackage{makecell}
\usepackage{array}
\usepackage{booktabs}
\usepackage{graphicx}
\usepackage{bm}
\usepackage{footnote}
\usepackage{cite}
\usepackage{threeparttable}
\usepackage[colorlinks,citecolor=blue]{hyperref}
\makesavenoteenv{tabular}
\makesavenoteenv{table}
\usepackage{array}

\begin{document}
%
\title{Revisiting Recursive Least Squares \\for  Training Deep Neural Networks}
%
%
%

\author{Chunyuan~Zhang,
        Qi~Song,
        Hui~Zhou,
        Yigui~Ou,
        Hongyao~Deng,
        and~Laurence Tianruo Yang, \textit{Fellow}, \textit{IEEE}
\thanks{This work was supported by the National Natural Science Foundation of China under Grant 61762032 and Grant 11961018. (Corresponding author: Chunyuan Zhang).}
\thanks{C. Zhang, Q. Song, H. Zhou  and L. T. Yang are with the School of Computer Science
and Technology, Hainan University, Haikou 570228, China (email: 990721@hainanu.edu.cn; 20151681310183@hainanu.edu.cn; zhouhui@hainanu.edu.cn;
ltyang@hainanu.edu.cn).}
\thanks{Y. Ou is with the School of Science,
Hainan University, Haikou 570228, China (email: 990661@hainanu.edu.cn).}
\thanks{H. Deng is with College of Computer Engineering, Yangtze Normal University, Chongqing 408100,
China (email: denghongyao@yznu.cn).}
\thanks{Manuscript received September 6, 2021.}
}

%
%

\markboth{IEEE TRANSACTIONS ON NEURAL NETWORKS AND LEARNING SYSTEMS}%
{Shell \MakeLowercase{\textit{et al.}}: Revisiting Recursive Least Squares for Training Deep Neural Networks}
%



\maketitle

\begin{abstract}
Recursive least squares (RLS) algorithms were once widely used for training small-scale neural networks,
due to their fast convergence. However, previous RLS algorithms are unsuitable for training deep neural
networks (DNNs), since they have high computational complexity and too many preconditions. In this paper,
to overcome these drawbacks, we propose three novel RLS optimization algorithms for training feedforward
neural networks, convolutional neural networks and recurrent neural networks (including long short-term
memory networks), by using the error backpropagation and our
average-approximation  RLS method, together with the equivalent gradients of the linear least squares
loss function with respect to the linear outputs of hidden layers. Compared with previous RLS optimization
algorithms, our algorithms are simple and elegant. They can be viewed as an improved stochastic gradient
descent (SGD) algorithm, which uses the inverse autocorrelation matrix of each layer as the adaptive learning rate.
Their time and space complexities are only several times those of SGD.
They only require the loss function to be the mean squared error and the activation function of the
output layer to be invertible. In fact, our algorithms can be also used in combination with other first-order
optimization algorithms without requiring these two preconditions. In addition,
we present two improved methods for our algorithms. Finally, we
demonstrate their effectiveness compared to the Adam algorithm on MNIST, CIFAR-10 and IMDB datasets,
and investigate the influences of their hyperparameters experimentally.
\end{abstract}

\IEEEpeerreviewmaketitle

\section{Introduction}
\IEEEPARstart{D}{eep} learning (DL) is an important branch of machine learning based
on deep neural networks (DNNs), such as feedforward neural networks (FNNs), convolutional neural
networks (CNNs), recurrent neural networks (RNNs) and long short-term memory networks (LSTMs) \cite{Goodfellow2016}. By using
multiple layers of linear or nonlinear processing neurons for
feature extraction and transformation, DL can directly learn representations of raw data with
multiple levels of abstraction \cite{LeCunBH2015}. Over the past decade, DL has succeeded in
many fields, including computer vision \cite{VoulodimosDDP2018}, speech
recognition \cite{Malik2021}, natural language processing \cite{Otter2021}, recommender system \cite{mu2018}, healthcare
 \cite{Esteva2019}, etc. However, there are still many challenges in DL. One of
the most difficult challenges is the parameter optimization problem of DNNs. During the
training process, the optimization often suffers from many difficulties, such as local
minima, saddle points, vanishing or exploding gradients, ill-conditioning, and long-term
dependencies \cite{Goodfellow2016}. It is quite common to take some days to weeks for
training a large-scale DNN in practice. With DNNs and datasets growing rapidly in size,
the  optimization problem has become a serious bottleneck blocking the widespread usage of DNNs,
and thus has received more and more research interest. Currently,
there are two main categories of optimization algorithms to address this problem,
that is, first-order and second-order algorithms.

First-order algorithms mainly use the first-order gradient information
to update the parameters of DNNs iteratively. Among these algorithms, stochastic gradient
descent (SGD) \cite{Bottou1991} is probably the most widely used one. At each iteration,
SGD randomly selects a minibatch from the training dataset and calculates the average gradient for training DNNs.
Although this learning mode makes SGD able to reduce the training time of DNNs,
especially when the training dataset is very large, SGD still has two main drawbacks. One drawback
is that the convergence of SGD is  slow. In order to tackle this issue,
several SGDs with momentum \cite{Sutskever2013} have been suggested. They accumulate
a decaying sum of the gradients into a momentum vector to update parameters. The
other drawback is that the learning rate of SGD is fixed and unrobust.
In the face of this, some modifications of SGD with adaptive learning rate have  been
developed. The most popular algorithms of this type mainly include AdaGrad \cite{Duchi2011},
RMSProp \cite{Tieleman2012} and Adam \cite{Kingma2015}. By accumulating the squared gradients
to design the adaptive learning rate, AdaGrad performs larger updates in the more gently
sloped directions. RMSProp can be viewed as an improved algorithm of AdaGrad. It replaces
the squared gradient accumulation with an exponentially weighted moving average, and is
more suitable for training DNNs. Adam is established on the adaptive estimates of first-order
and second-order moments, which employ momentums and the moving average squared gradients,
respectively. Therefore, Adam can also be viewed as a hybrid of RMSProp and SGDs with momentum.
In practice, these modified algorithms have been proven more effective than SGD. However,
they sometimes lead to worse generalization performance \cite{Wilson2017}.

Second-order algorithms typically use the second-order derivatives of the
loss function to speed up the training of DNNs \cite{martens2016}. The best known second-order algorithm
should be the Newton's algorithm \cite{Battiti1992}. By using the curvature information, in the
form of the Hessian, to rescale the gradients, it can jump to the minimum in far fewer steps than
first-order algorithms. However, it has two severe problems for training DNNs. One
problem is that the Hessian is often not positive definite, since
the loss function of DNNs is generally nonconvex. This problem is often tackled by means of
regularization methods \cite{Goodfellow2016}. The other problem is that it
requires computing the Hessian and the inverse Hessian at each update.
The recent advances for this problem can be roughly categorized into Hessian-free
algorithms, stochastic Gaussian-Newton algorithms, stochastic quasi-Newton algorithms and
trust region algorithms \cite{Chen2018}. The first type often uses the finite differences method
to compute  Hessian-vector products, and then uses the linear conjugate gradient
method to obtain the search direction \cite{Martens2010,Martens2011}.
The second type, such as the K-FAC algorithm \cite{Martens2015} and
the K-FRA algorithm \cite{Botev2017}, uses the block-diagonal approximation of the Hessian to
reduce the computational cost. The third type mainly uses
the L-BFGS method to approximate the inverse Hessian \cite{Wang2017,Goldfarb2020}.
The last type determines the search direction by minimizing the quadratic approximation
 around the current solution with the norm constraints \cite{Xu2020}.
These algorithms have been proven more effective than first-order algorithms.
Nevertheless, they mostly require tens to hundreds of iterations at each update, and thus are still expensive.
In addition, most of them can only be used for training some specific types of DNNs.

The above algorithms are based on nonlinear optimization techniques, which
generally have some intrinsic drawbacks such as slow convergence or intensive
computation. As we know, the computational model of an artificial neuron can be decomposed into a linear
affine part and a nonlinear activation part. Intuitively, if we can calculate the output loss
of the linear affine part, we will be able to employ linear optimization
techniques for training DNNs. In fact, in the era of neural network connectionism,
there had been many linear optimization algorithms, which mostly used the linear
output loss to replace the activation output loss for backward propagation,
and used the linear or recursive least squares methods to update the parameters of neural networks
layer by layer \cite{Yam1995, Ergezinger1995, Yam1997, Biegler-Konig1993,Fontenla-Romero2003, Azimi-Sadjadi1990, Azimi-Sadjadi1992, Cho2001, Al-Batah2010, Stan2000, Bilski1998}. Compared with the steepest gradient descent algorithm, they
exhibit extremely fast convergence. However, they are unsuitable for training DNNs, since they have many drawbacks.
Firstly, they require the activation function of each layer to be invertible and even to be piecewise linear \cite{Azimi-Sadjadi1990,Azimi-Sadjadi1992}.
Secondly, they use the individual sample learning rather than the minibatch learning, and thus can't scale to large datasets.
Thirdly, most of them are designed for training FNNs, and only a few  are designed for training RNNs \cite{Cho2001}.
Lastly, most of them have high computational and space complexity, and some of them even require preserving
one inverse autocorrelation matrix for each neuron \cite{Azimi-Sadjadi1990, Stan2000, Bilski1998}.
As a result, the linear or recursive least squares methods seem to have been forgotten in DL. Whereas, in striking contrast, they are full of vigor
and vitality in extreme learning machines, which are a special type of neural networks without training hidden neurons \cite{Huang2006, Huang2012, Zhou2015, Pang2016, Park2017}. Perhaps it is time to revisit them for training DNNs.

In this paper, we focus on the recursive least squares (RLS) method for training DNNs, since it is more suitable
for online learning than the linear least squares method. We propose three stochastic RLS optimization algorithms
for training FNNs, CNNs and RNNs (including LSTMs), respectively.
Our contributions mainly include the following five aspects:
1) Our algorithms use the popular minibatch mode for training DNNs. In order to achieve this, we
present an average-approximation RLS method, which uses the average values of minibatch inputs to update the inverse
autocorrelation matrix of each layer.
2) Our algorithms provide a novel equivalent method to define the gradients of the least squares loss
function with respect to (w.r.t.)  the linear outputs of hidden layers. By using the equivalent gradients, we can easily derive
the least squares solutions of the parameters of each hidden layer, without requiring the activation
function of any hidden layer to be invertible.
3) Our algorithms are simple and easy to implement. We convert them into a special form of SGD,
which uses the mean squared error (MSE) between actual linear outputs and desired linear outputs as the loss function and uses
the inverse autocorrelation matrix of each layer as the adaptive learning rate.
4) Our algorithms have fast convergence and low time complexity.
Their convergence performance is superior to that of Adam, and their time and space complexities are only several times
those of the conventional SGD algorithm.
5) Lastly, we discuss some possible improvements and present two improved methods for our algorithms.

The rest of this paper is organized as follows.
In Section II, we review the mathematical models of the four most popular types of DNNs and introduce some notations.
In Section III, we introduce the derivation of the RLS optimization for FNNs in detail.
In Section IV, we also introduce the derivation of  RLS optimization for CNNs and RNNs, respectively.
In Section V, we analyze the complexities of our proposed algorithms,
and present two improved methods.
In Section VI,  we verify the effectiveness of our proposed algorithms through three experiments, and investigate
the influences of hyperparameters on the performances of our proposed algorithms.
Finally, conclusions are summarized in Section VII.

\section{Deep Neural Networks}
In this section, for facilitating the derivation of our proposed algorithms,
we briefly review the mathematical models of FNNs, CNNs, RNNs and LSTMs, respectively. In addition,
we introduce some notations used in this paper.
\subsection{Feedforward Neural Networks}
\begin{figure*}[htbp]
\centering
\includegraphics[scale=0.96,trim=0 0 0 0]{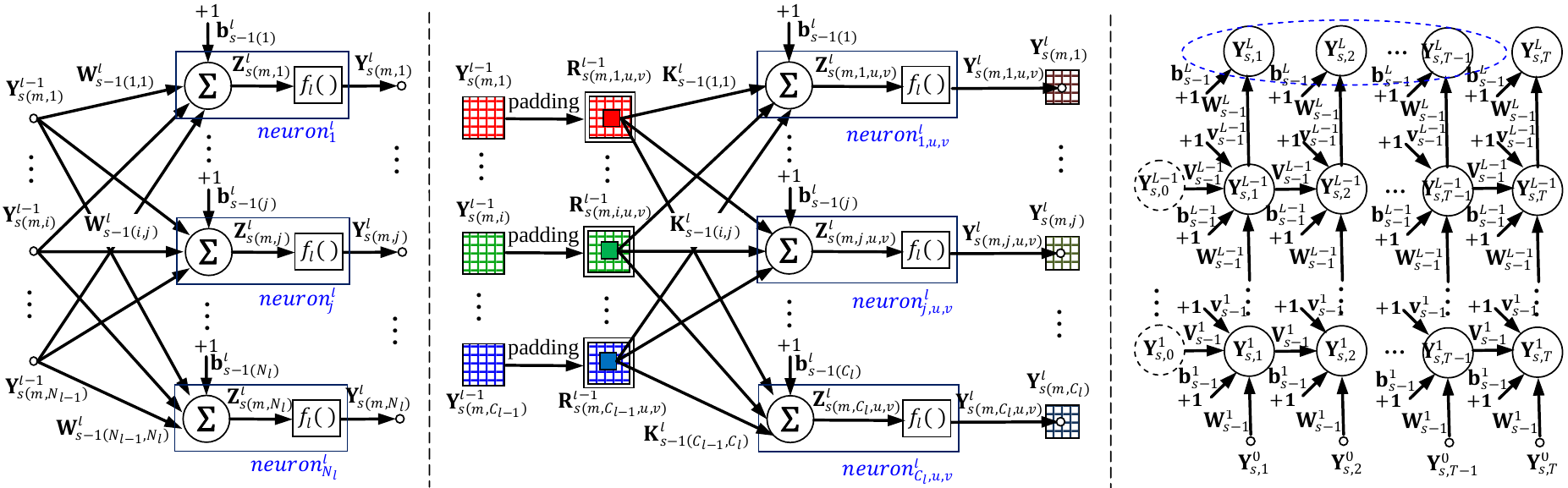}\\
\footnotesize{\hspace{0.2cm}(a)\hspace{6.1cm}(b)\hspace{6.4cm}(c)}
\caption{(a) Structure of a fully-connected layer in an FNN. (b) Structure of a convolutional layer in a CNN. (c) Unfolded structure of a stacked RNN.}
\end{figure*}
An FNN generally consists of several fully-connected (FC) layers.
In this paper, we consider all types of DNNs with $L$ layers, wherein the $0^{\textrm{th}}$ layer is the input
layer, and the $L^{\textrm{th}}$ layer is the output layer.
Let $N_l$, $\textbf{W}_{s-1}^{l} \in \mathbb{R}^{N_{l-1}\times N_{l}}$, $\textbf{b}_{s-1}^{l} \in \mathbb{R}^ {N_{l}}$ and $f_l(\cdot)$ denote the number of neurons, the weight matrix, the bias vector and
the activation function of the $l^{\textrm{th}}$ layer in an FNN at step $s$, respectively.
Give a minibatch training dataset $\{(\textbf{Y}_{s}^{0},\textbf{Y}_{s}^{*})|\textbf{Y}_{s}^{0} \in \mathbb{R}^{M\times N_{0}},\textbf{Y}_{s}^{*} \in \mathbb{R}^{M\times N_{L}}\}_{s\geqslant1}$, where $\textbf{Y}_{s}^{0}$ is the  input matrix of the FNN at step $s$,
$\textbf{Y}_{s}^{*}$ is the corresponding desired output matrix, and $M$ is the minibatch size.
Then, the structure of an FC layer in an FNN can be illustrated in Fig. 1(a), where
$neuron^l_j$ denotes the $j^{\textrm{th}}$ neuron,
$\textbf{Y}_{s}^{l-1} \in \mathbb{R}^{M\times N_{l-1}}$,
$\textbf{Z}_{s}^{l} \in \mathbb{R}^{M \times N_{l}}$ and
$\textbf{Y}_{s}^{l} \in \mathbb{R}^{M\times N_{l}}$ are the matrices of the inputs from the preceding layer, the actual linear outputs and the
actual activation outputs, respectively.
Note that the above notations also have the same or similar meanings when they are used in
CNNs, RNNs and LSTMs.

At each step, FNNs usually use the forward propagation of inputs to calculate actual activation outputs, and then use the
backward propagation of errors to update parameters \cite{LeCunBH2015}. From Fig. 1(a), the forward propagation of each layer
can be described as
\begin{eqnarray}
\textbf{Y}_{s}^{l}= f_{l}(\textbf{Z}_{s}^{l}) = f_{l}(\textbf{Y}_{s}^{l-1}\textbf{W}_{s-1}^{l}+\textbf{1}\times(\textbf{b}_{s-1}^{l})^\textrm{T})
\end{eqnarray}
where  $l=1,\cdots,L$, $\textbf{1} \in \mathbb{R}^{M}$
is the all-ones vector,  $\textrm{T}$  is the transposition,
and $\textbf{Z}_{s}^{l} = \textbf{Y}_{s}^{l-1}\textbf{W}_{s-1}^{l}+\textbf{1}\times(\textbf{b}_{s-1}^{l})^\textrm{T}$.
For brevity and clarity, we use the augmented matrices to rewrite  (1)  as
\begin{eqnarray}
\textbf{Y}_{s}^{l}=f_{l}(\textbf{Z}_{s}^{l})=f_l(\textbf{X}_{s}^{l}\mathbf{\Theta}_{s-1}^{l})
\end{eqnarray}
where $\textbf{X}_{s}^l = [\textbf{Y}_{s}^{l-1},\textbf{1}]$,
$\mathbf{\Theta}_{s-1}^l = [(\textbf{W}_{s-1}^{l})^\textrm{T}, {\textbf{b}_{s-1}^{l}}]^\textrm{T}$
and $\textbf{Z}_{s}^{l} = \textbf{X}_{s}^{l}\mathbf{\Theta}_{s-1}^{l} $.
After forward propagation layer by layer, FNNs  backward propagate the
errors between $\textbf{Y}_{s}^{L}$  and $\textbf{Y}_{s}^{*}$,
and update $\{\mathbf{\Theta}_{s-1}^{l}\}_{l=1}^{L}$ to  $\{\mathbf{\Theta}_{s}^{l}\}_{l=1}^{L}$ with an optimization algorithm.

\subsection{Convolutional Neural Networks}
A CNN generally consists of some convolutional (CONV), pooling and FC layers.
Since FC layers in CNNs have the same learning mode as in FNNs and
pooling layers usually has no learnable parameters,
we only review the mathematical model of  CONV
layers in this subsection. Suppose that all CONV layers perform the two-dimensional convolution.
Let $C_l$, $U_l$, $V_l$, $H_l$, $W_l$, $\textbf{K}_{s-1}^{l} \in \mathbb{R}^{C_{l-1}\times C_{l} \times H_l \times W_l}$ and $\textbf{b}_{s-1}^{l} \in \mathbb{R}^ {C_{l}}$ denote the number of output channels, the height of output channels,
the width of output channels, the filter (or kernel) height, the filter width, the filter tensor and
the bias vector of a CONV layer at step $s$, respectively. Give a minibatch training dataset $\{(\textbf{Y}_{s}^{0},\textbf{Y}_{s}^{*})|\textbf{Y}_{s}^{0} \in \mathbb{R}^{M\times C_{0}\times U_0 \times V_0},\textbf{Y}_{s}^{*} \in \mathbb{R}^{M\times N_{L}}\}_{s\geqslant1}$, where $\textbf{Y}_{s}^{0}$ is the input tensor of the CNN at step $s$, and $\textbf{Y}_{s}^{*}$ is the corresponding desired output matrix.
Then, the structure of a CONV layer can be illustrated in Fig. 1(b), where $neuron^l_{j,u,v}$ denotes the
neuron in the $u^{\textrm{th}}$ row and the $v^{\textrm{th}}$ column of the $j^{\textrm{th}}$ output channel,
$\textbf{Y}_{s}^{l-1} \in \mathbb{R}^{M\times C_{l-1}\times U_{l-1}\times V_{l-1}}$,
$\textbf{Z}_{s}^{l} \in \mathbb{R}^{M \times C_{l}\times U_{l}\times V_{l}}$ and
$\textbf{Y}_{s}^{l} \in \mathbb{R}^{M\times C_{l}\times U_{l}\times V_{l}}$ are the tensors of the inputs, the actual linear outputs and the
actual activation outputs, respectively,
and $\textbf{R}^{l-1}_{{s}(m,i,u,v)} \in  \mathbb{R}^{H_l \times W_{l}}$ denotes the receptive field in the $i^{\textrm{th}}$ input channel to $neuron^l_{\cdot,u,v}$.

From Fig. 1(b), $\textbf{R}^{l-1}_{{s}(\cdot,\cdot,u,v)}$ is fully connected to $neuron^l_{:,u,v}$, where ``$:$" denotes
all elements in the specified dimension.
Thus, for derivational convenience in Section \uppercase\expandafter{\romannumeral4}.\textit{A}, we formulate the forward propagation of the CONV layer in a form similar to that of the FC layer.
According to the definition of the CONV operation \cite{Goodfellow2016},
$\textbf{R}^{l-1}_{s(:,:,u,v,h,w)}$  is defined as
\begin{eqnarray}
\textbf{R}^{l-1}_{s(:,:,u,v,h,w)}
 = \textbf{\textbf{Y}}^{l-1}_{s(:,:,(u-1)\times D_{l}+h-P_{l},(v-1)\times D_{l}+w-P_{l})}
\end{eqnarray}
where $D_l$ and  $P_l$ are the stride and the padding size in the $l^{\textrm{th}}$ layer.
Reshape
$\textbf{R}^{l-1}_{s}$ as $\textbf{\~{R}}^{l-1}_{s} \in \mathbb{R}^{M \times C_{l-1}H_lW_l\times U_{l} \times V_{l}}$,
where  $\textbf{\~{R}}^{l-1}_{s(m,:,u,v)} \in \mathbb{R}^{C_{l-1}H_lW_l}$ is defined as
\begin{eqnarray}
\textbf{\~{R}}^{l-1}_{s(m,:,u,v)}
=\textrm{flatten}(\textbf{R}^{l-1}_{s(m,:,u,v)})
\end{eqnarray}
where $\textrm{flatten}(\cdot)$ denotes reshaping the given tensor into a column vector.
Similarly, reshape
$\textbf{K}^{l}_{s-1}$ as $\textbf{\~{K}}^{l}_{s-1} \in \mathbb{R}^{ C_{l-1}H_lW_l\times C_{l}}$,
where $\textbf{\~{K}}^{l}_{s-1(:,j)} \in \mathbb{R}^{C_{l-1}H_lW_l}$ is defined as
\begin{eqnarray}
\textbf{\~{K}}^{l}_{s-1(:,j)}
= \textrm{flatten}(\textbf{K}^{l}_{s-1(:,j)})
\end{eqnarray}
Then, the actual activation outputs of each CONV layer are calculated as
\begin{eqnarray}
\textbf{Y}_{s(:,:,u,v)}^{l}= f_{l}(\textbf{Z}_{s(:,:,u,v)}^{l}) = f_{l}(\textbf{X}_{s(:,:,u,v)}^{l}\mathbf{\Theta}_{s-1}^{l})
\end{eqnarray}
where $\textbf{X}_{s(:,:,u,v)}^l = [\textbf{\~R}_{s(:,:,u,v)}^{l-1},\textbf{1}]$,
$\mathbf{\Theta}_{s-1}^l = [(\textbf{\~K}_{s-1}^{l})^\textrm{T}, {\textbf{b}_{s-1}^{l}}]^\textrm{T}$
and $\textbf{Z}_{s(:,:,u,v)}^{l}
= \textbf{X}_{s(:,:,u,v)}^{l}\mathbf{\Theta}_{s-1}^{l} $. $\mathbf{\Theta}_{s-1}^{l}$ is generally
updated to $\mathbf{\Theta}_{s}^{l}$ by using the backward propagation of errors as well.

\subsection{Recurrent Neural Networks}
A stacked RNN generally consists of several recurrent (RECUR) layers and one FC layer.
Unlike FNNs and CNNs,
RNNs are mainly used for processing sequential data.
Let $N_l$, $\textbf{W}_{s-1}^{l} \in \mathbb{R}^{N_{l-1}\times N_{l}}$, $\textbf{b}_{s-1}^{l} \in \mathbb{R}^{N_{l}}$, $\textbf{V}_{s-1}^{l} \in \mathbb{R}^{N_{l}\times N_{l}}$ and $\textbf{v}_{s-1}^{l} \in \mathbb{R}^{N_l}$ denote the number of neurons, the FC weight matrix, the FC bias vector, the RECUR weight matrix and the RECUR
bias vector.
Note here that we use two bias vectors for facilitating the implementation of our proposed algorithm in Section \uppercase\expandafter{\romannumeral4}.\textit{B} by using PyTorch.
Give a minibatch training dataset $\{(\{\textbf{Y}_{s,t}^{0}\}_{t=1}^T,\{\textbf{Y}_{s,t}^{*}\}_{t=t_0}^{T})| \textbf{Y}_{s,t}^{0} \in \mathbb{R}^{M\times N_{0}}, \textbf{Y}_{s,t}^{*} \in \mathbb{R}^{M\times N_{L}}\}_{s\geqslant1}$, where $\textbf{Y}_{s,t}^{0} $ is the input matrix of the RNN at time $t$ in step $s$, $\textbf{Y}_{s,t}^{*} \in \mathbb{R}^{M\times N_{L}}$ is the corresponding desired activation output matrix, $T$ is the sequence length,
and $t_0$ is $1$ for sequence prediction problems or $T$ for sequence classification problems.
Then, the unfolded structure of a stacked RNN is illustrated in Fig. 1(c), where the outputs in the blue dashed ellipse denote that they are
not required for sequence classification problems,
$\textbf{W}_{s-1}^{l}$, $\textbf{b}_{s-1}^{l}$, $\textbf{V}_{s-1}^{l}$ and $\textbf{v}_{s-1}^{l}$
are shared across the $s^{\textrm{th}}$ minibatch sequence, and $\textbf{Y}_{s,0}^{l}$ is the initial RECUR input matrix at time 0.

Different from FNNs and CNNs, RNNs usually use the forward propagation through time algorithm to
calculate actual activation outputs. From Fig. 1(c), at time $t$, the outputs of each RECUR layer are calculated as
\begin{eqnarray}
\textbf{Y}_{s, t}^{l}=f_l(\textbf{Z}^{l_w}_{s, t}+\textbf{Z}^{l_v}_{s, t})=f_{l}(\textbf{X}_{s,t}^{l_w}\mathbf{\Theta}_{s-1}^{l_w}+\textbf{X}_{s,t}^{l_v}\mathbf{\Theta}_{s-1}^{l_v})
\end{eqnarray}
where $t=1,\cdots,T$, $l=1,\cdots,L-1$,
$\textbf{X}_{s,t}^{l_w} = [\textbf{Y}_{s,t}^{l-1}, \textbf{1}]$,
$\mathbf{\Theta}_{s-1}^{l_w} = [(\textbf{W}_{s-1}^{l})^\textrm{T}, {\textbf{b}_{s-1}^{l}}]^\textrm{T}$,
$\textbf{Z}_{s,t}^{l_w} = \textbf{X}_{s,t}^{l_w}\mathbf{\Theta}_{s-1}^{l_w}$,
$\textbf{X}_{s,t}^{l_v} = [\textbf{Y}_{s,t-1}^{l}, \textbf{1}]$,
$\mathbf{\Theta}_{s-1}^{l_v} = [(\textbf{V}_{s-1}^{l})^\textrm{T}, {\textbf{v}_{s-1}^l}]^\textrm{T}$
and $\textbf{Z}_{s,t}^{l_v} = \textbf{X}_{s,t}^{l_v}\mathbf{\Theta}_{s-1}^{l_v}$.
The  outputs of the output layer are calculated as
\begin{eqnarray}
\textbf{Y}_{s, t}^{L}=f_L(\textbf{Z}^{L}_{s,t})=f_{L}(\textbf{X}_{s,t}^{L}\mathbf{\Theta}_{s-1}^{L})
\end{eqnarray}
where $t=t_0,\cdots,T$, $\mathbf{\Theta}_{s-1}^{L} = [(\textbf{W}_{s-1}^{L})^\textrm{T}, \textbf{b}_{s-1}^{L}]^\textrm{T}$,
$\textbf{X}_{s,t}^{L} = [\textbf{Y}_{s,t}^{l-1}, \textbf{1}]$ and
$\textbf{Z}_{s,t}^{L} = \textbf{X}_{s,t}^{L}\mathbf{\Theta}_{s-1}^{L}$.
After forward propagation through time,  RNNs generally use the backward propagation through time (BPTT)
method  \cite{Jaeger2002} to update $\mathbf{\Theta}_{s-1}^L$
and $\{(\mathbf{\Theta}_{s-1}^{l_w}, \mathbf{\Theta}_{s-1}^{l_v})\}_{l=1}^{L-1}$ with an optimization algorithm.

\subsection{LSTM Neural Networks}
An LSTM is a special case of RNNs. It has the same input and output layers as ordinary RNNs,
but uses LSTM units instead of ordinary RECUR neurons in hidden layers (called LSTM layers).
Therefore, we only review the mathematical model of stacked LSTM layers in this subsection. An LSTM unit is composed of an input gate,
an input modulation gate, a forget gate, a cell and an output gate \cite{Hochreiter1997}. All gates have same inputs, but have different parameters.
Let
$\textbf{W}_{s-1}^{l_{(\cdot)}} \in \mathbb{R}^{N_{l-1}\times N_{l}}$,  $\textbf{b}_{s-1}^{l_{(\cdot)}}
\in \mathbb{R}^{N_l}$, $\textbf{V}_{s-1}^{l_{(\cdot)}} \in \mathbb{R}^{N_{l}\times N_{l}}$  and $\textbf{v}_{s-1}^{l_{(\cdot)}}
\in \mathbb{R}^{N_l}$ denote the FC weight matrix, the FC bias vector, the RECUR weight matrix and the RECUR bias vector for a type of gate $(\cdot)$ in an
LSTM layer. Here, we use $\mathrm{(I)}$, $\mathrm{(G)}$,$\mathrm{(F)}$ and $\mathrm{(O)}$ to denote the input,
input modulation, forget and output gates, respectively.

Similar to a RECUR layer, an LSTM layer also uses the forward propagation through time to calculate the actual activation
outputs. In order to make LSTMs able to use our proposed algorithm in Section \uppercase\expandafter{\romannumeral4}.\textit{B} directly, we concatenate all parameters of an LSTM layer as follows
\begin{eqnarray}
\textbf{W}_{s-1}^{l} = [\textbf{W}_{s-1}^{l_{(\mathrm{I})}}, \textbf{W}_{s-1}^{l_{(\mathrm{G})}}, \textbf{W}_{s-1}^{l_{(\mathrm{F})}},
\textbf{W}_{s-1}^{l_{(\mathrm{O})}}]~~~~~
\\
\textbf{b}_{s-1}^{l} = [(\textbf{b}_{s-1}^{l_{(\mathrm{I})}})^\textrm{T}, (\textbf{b}_{s-1}^{l_{(\mathrm{G})}})^\textrm{T}, (\textbf{b}_{s-1}^{l_{(\mathrm{F})}})^\textrm{T},(\textbf{b}_{s-1}^{l_{(\mathrm{O})}})^\textrm{T}]^\textrm{T}
\end{eqnarray}
\begin{eqnarray}
\textbf{V}_{s-1}^{l} = [\textbf{V}_{s-1}^{l_{(\mathrm{I})}}, \textbf{V}_{s-1}^{l_{(\mathrm{G})}}, \textbf{V}_{s-1}^{l_{(\mathrm{F})}}, \textbf{V}_{s-1}^{l_{(\mathrm{O})}}]~~~~~~\\
\textbf{v}_{s-1}^{l} = [(\textbf{v}_{s-1}^{l_{(\mathrm{I})}})^\textrm{T}, (\textbf{v}_{s-1}^{l_{(\mathrm{G})}})^\textrm{T}, (\textbf{v}_{s-1}^{l_{(\mathrm{F})}})^\textrm{T},(\textbf{v}_{s-1}^{l_{(\mathrm{O})}})^\textrm{T}]^\textrm{T}
\end{eqnarray}
Then, at time $t$, for the same training dataset given in Section \uppercase\expandafter{\romannumeral2}.\textit{C},
we can also define $\textbf{Z}_{s,t}^{l_w} = \textbf{X}_{s,t}^{l_w}\mathbf{\Theta}_{s-1}^{l_w}$
and $\textbf{Z}_{s,t}^{l_v} = \textbf{X}_{s,t}^{l_v}\mathbf{\Theta}_{s-1}^{l_v}$, where
$\textbf{X}_{s,t}^{l_w}$,
$\mathbf{\Theta}_{s-1}^{l_w}$,
$\textbf{X}_{s,t}^{l_v}$ and
$\mathbf{\Theta}_{s-1}^{l_v}$
have the same forms defined in Section \uppercase\expandafter{\romannumeral2}.\textit{C}.
On this basis, four gate outputs of this layer can be described as follows
\begin{eqnarray}
\textbf{Y}_{s, t}^{l_{\mathrm{(I)}}} = \sigma(\textbf{Z}^{l_w}_{s, t(:,1:N_l)}+\textbf{Z}^{l_v}_{s, t(:,1:N_l)})~~~~~~\\
\textbf{Y}_{s, t}^{l_{\mathrm{(G)}}} = g(\textbf{Z}^{l_w}_{s, t(:,N_l+1:2N_l)}+\textbf{Z}^{l_v}_{s, t(:,N_l+1:2N_l)})~\\
\textbf{Y}_{s, t}^{l_{\mathrm{(F)}}} = \sigma(\textbf{Z}^{l_w}_{s, t(:,2N_l+1:3N_l)}+\textbf{Z}^{l_v}_{s, t(:,2N_l+1:3N_l)})\\
\textbf{Y}_{s, t}^{l_{\mathrm{(O)}}} = \sigma(\textbf{Z}^{l_w}_{s, t(:,3N_l+1:4N_l)}+\textbf{Z}^{l_v}_{s, t(:,3N_l+1:4N_l)})
\end{eqnarray}
where $\sigma(\cdot)$ and $g(\cdot)$  denotes the activation functions, and they commonly set to sigmoid($\cdot$) and tanh($\cdot$), respectively.
 The cell outputs $\textbf{C}_{s, t}^{l}$ and the actual activation outputs $\textbf{Y}_{s, t}^{l}$ of this layer are described as follows
\begin{eqnarray}
\textbf{C}_{s, t}^{l}= \textbf{Y}_{s, t}^{l_{\mathrm{(I)}}} \circ \textbf{Y}_{s, t}^{l_{\mathrm{(G)}}} + \textbf{Y}_{s, t}^{l_{\mathrm{(F)}}}\circ\textbf{C}_{s, t-1}^{l}\\
\textbf{Y}_{s, t}^{l}=\textbf{Y}_{s, t}^{l_{\mathrm{(O)}}} \circ g(\textbf{C}_{s,t}^{l})~~~~~~~~
\end{eqnarray}
where $\circ$ is the Hadamard product. Generally, the update of $\{(\mathbf{\Theta}_{s-1}^{l_w}, \mathbf{\Theta}_{s-1}^{l_v})\}_{l=1}^{L-1}$
is the same as that in RNNs.

\section{RLS Optimization for FNNs}
In this section, we introduce the derivation of the RLS optimization for FNNs in detail. We first
define a linear least squares loss function for FNNs. Next, we provide the derivations of average-approximation RLS formulas for optimizing
output and hidden layers, respectively. Finally, we present the pseudocode of the practical implementation.

\subsection{Defining Loss Function}
In order to convert the FNN optimization into the linear least squares problem, we
assume that the activation function of the output layer is invertible.
Considering the minibatch training dataset $\{(\textbf{Y}_{i}^{0},\textbf{Y}_{i}^{*})\}_{i=1}^{s}$,
we define the loss function by using the linear output errors as
\begin{eqnarray}
J(\mathbf{\Theta})=\frac{1}{2M}\sum_{i=1}^s \lambda^{s-i} \left\|\textbf{Z}_{i}^{L}-\textbf{Z}_{i}^{L*}\right\|_\textsc{F}^2
\end{eqnarray}
where $\mathbf{\Theta}$ denotes all augmented parameter matrices of the FNN,
$\mathbf{Z}_{i}^{L*}=f_L^{-1}(\textbf{Y}_{i}^*)$, $\mathbf{Z}_{i}^{L} = \textbf{X}_{i}^{L}\mathbf{\Theta}^{L}$,
and $\lambda \in (0,1]$ is the forgetting factor.
Then, the FNN optimization problem can be formally described as
\begin{eqnarray}
\mathbf{\Theta}_s = \mathop\textrm{argmin}_{\mathbf{\Theta}} J(\mathbf{\Theta})
\end{eqnarray}
\subsection{Optimizing Output Layer}
Firstly, we try to derive the least squares solution of $\mathbf{\Theta}_s^L$.
Let ${\mathbf{\nabla}}_{\mathbf{\Theta}^L}$ denote $\partial  J(\mathbf{\Theta})/\partial \mathbf{\Theta}^L$.
Using  (19) and the chain rule for $\mathbf{\Theta}^L$, we easily have
\begin{eqnarray}
{\mathbf{\nabla}}_{\mathbf{\Theta}^L} =\sum_{i=1}^s (\textbf{X}_{i}^L)^\textrm{T} {\mathbf{\nabla}}_{\textbf{Z}_{i}^L}
\end{eqnarray}
where ${\mathbf{\nabla}}_{\textbf{Z}_{i}^L}$ denotes $\partial  J(\mathbf{\Theta})/\partial \textbf{Z}_{i}^L$, which is defined as
\begin{eqnarray}
{\mathbf{\nabla}}_{\textbf{Z}_{i}^L}=\frac{\lambda^{s-i}}{M}(\textbf{Z}_{i}^{L}-\textbf{Z}_{i}^{L*})
\end{eqnarray}
Let ${\mathbf{\nabla}}_{\mathbf{\Theta}^L}=\textbf{0}$ such that
\begin{eqnarray}
\sum_{i=1}^s (\textbf{X}_{i}^L)^\textrm{T} {\mathbf{\nabla}}_{\textbf{Z}_{i}^L} =\textbf{0}
\end{eqnarray}
Plugging  (22) into (23), we can obtain
\begin{eqnarray}
\mathbf{\Theta}_s^L =(\textbf{A}_s^L)^{-1} \textbf{B}_s^L
\end{eqnarray}
where $\textbf{A}_s^L$ and $\textbf{B}_s^L$ are defined as follows
\begin{eqnarray}
\textbf{A}_s^L = \frac{1}{M} \sum_{i=1}^s \lambda^{s-i}(\textbf{X}_{i}^L)^\textrm{T} \textbf{X}_{i}^L ~   \\
\textbf{B}_s^L =  \frac{1}{M} \sum_{i=1}^s \lambda^{s-i}(\textbf{X}_{i}^L)^\textrm{T} \textbf{Z}_{i}^{L*}
\end{eqnarray}

Secondly, in order to avoid computing the inverse of $\mathbf{A}_s^L$  and make our algorithm more suitable for online learning,
we try to derive the recursive update formulas of  $(\mathbf{A}_s^L)^{-1}$ and $\mathbf{\Theta}_s^L$. We rewrite
(25) and (26) as follows
\begin{eqnarray}
\textbf{A}_s^L= \lambda\textbf{A}_{s-1}^L +  \frac{1}{M} (\textbf{X}_{s}^L)^\textrm{T} \textbf{X}_{s}^L ~ \\
\textbf{B}_s^L= \lambda\textbf{B}_{s-1}^L + \frac{1}{M}  (\textbf{X}_{s}^L)^\textrm{T} \textbf{Z}_{s}^{L*}
\end{eqnarray}
Let $\textbf{P}_s^L =(\textbf{A}_s^L)^{-1}$ and use the Kailath variant formula \cite{Petersen2012} for (25).
Then, $\textbf{P}_s^L$ is recursively calculated by
$\textbf{P}_s^L = \frac{1}{\lambda} \textbf{P}_{s-1}^L - \frac{1}{\lambda} \textbf{P}_{s-1}^L (\textbf{X}_{s}^L)^\textrm{T}(\lambda M \textbf{I} +\textbf{X}_{s}^L \textbf{P}_{s-1}^L(\textbf{X}_{s}^L)^\textrm{T})^{-1}\textbf{X}_{s}^L \textbf{P}_{s-1}^L$,
where $\textbf{I} \in \mathbb{R}^{M \times M}$ is the identity matrix. Unfortunately, $\textbf{P}_s^L$ requires
computing the inverse of $ \lambda M \textbf{I} +\textbf{X}_{s}^L \textbf{P}_{s-1}^L(\textbf{X}_{s}^L)^\textrm{T}$.
As in the derivation of ordinary RLS, we rewrite (27) and (28) as the following vector-product forms
\begin{eqnarray}
\textbf{A}_s^L= \lambda\textbf{A}_{s-1}^L +  \frac{1}{M}  \sum_{m=1}^M\textbf{X}_{s(m,:)}^L(\textbf{X}_{s(m,:)}^L)^{\textrm{T}}  \\
\textbf{B}_s^L= \lambda\textbf{B}_{s-1}^L + \frac{1}{M}   \sum_{m=1}^M\textbf{X}_{s(m,:)}^L(\textbf{Z}_{s(m,:)}^{L*})^{\textrm{T}}
\end{eqnarray}
where $\textbf{X}_{s(m,:)}^L \in \mathbb{R}^{N_{L}}$ and $\textbf{Z}_{s(m,:)}^{L*} \in \mathbb{R}^{N_{L}}$
are the column vectors sliced from $\textbf{X}_{s}^L$ and $\textbf{Z}_{s}^{L*}$, respectively. Whereas,
we cannot use the Sherman-Morrison formula \cite{Petersen2012} for (29) to derive the recursive formula of $\textbf{P}_s^L$,
since the rightmost term in (29) is the average of vector products.

In order to tackle this problem, we present an average-approximation method, which uses the
average-vector product to replace the average of vector products approximately.
Define the average vectors of $\textbf{X}_{s}^L$ and $\textbf{Z}_{s}^{L*}$ as follows
\begin{eqnarray}
{\bm{\bar{x}}}_{s}^L  =\frac{1}{M} \sum_{m=1}^M (\textbf{X}_{s(m,:)}^L)^{\textrm{T}} & \\
{\bm{\bar{z}}}_{s}^{L*} =\frac{1}{M} \sum_{m=1}^M (\textbf{Z}_{s(m,:)}^{L*})^{\textrm{T}}
\end{eqnarray}
Suppose that there exists a ratio factor $k>0$ such that
\begin{eqnarray}
k{\bm{\bar{x}}}_{s}^L(\bm{\bar{x}}_{s}^L)^\textrm{T} \approx  \frac{1}{M}\sum_{m=1}^M\textbf{X}_{s(m,:)}^L(\textbf{X}_{s(m,:)}^L)^{\textrm{T}}= \frac{1}{M} (\textbf{X}_{s}^L)^\textrm{T} \textbf{X}_{s}^L ~ \\
k{\bm{\bar{x}}}_{s}^L(\bm{\bar{z}}^{L*}_{s})^\textrm{T} \approx \frac{1}{M}\sum_{m=1}^M\textbf{X}_{s(m,:)}^L(\textbf{Z}_{s(m,:)}^{L*})^{\textrm{T}}
= \frac{1}{M}  (\textbf{X}_{s}^L)^\textrm{T} \textbf{Z}_{s}^{L*}
\end{eqnarray}
Then, we rewrite (29) and (30) as follows
\begin{eqnarray}
\textbf{A}_{s}^L \approx \lambda\textbf{A}_{s-1}^L + k{\bm{\bar{x}}}_{s}^L(\bm{\bar{x}}_{s}^L)^\textrm{T}~ \\
\textbf{B}_{s}^L \approx \lambda\textbf{B}_{s-1}^L + k{\bm{\bar{x}}}_{s}^L(\bm{\bar{z}}^{L*}_{s})^\textrm{T}
\end{eqnarray}
Now, by using the Sherman-Morrison formula for (35), we can easily derive
\begin{eqnarray}
\textbf{P}_s^L \approx \frac{1}{\lambda} \textbf{P}_{s-1}^L -  \frac{k}{\lambda h_s^L} \bm{u}_{s}^L (\bm{u}_{s}^L)^{\textrm{T}}
\end{eqnarray}
where $\bm{u}_s^L$ and $h_s^L$ are defined as
\begin{eqnarray}
\bm{u}_s^L=\textbf{P}_{s-1}^L \bm{\bar{x}}_{s}^L  ~~~~&\\
h_{s}^L=\lambda+ k(\bm{\bar{x}}_{s}^L)^{\textrm{T}}\bm{u}_s^L
\end{eqnarray}
Then, plugging  (36) and (37) into  (24), we can get
\begin{eqnarray}
\mathbf{\Theta}_s^L \approx \mathbf{\Theta}_{s-1}^L - \frac{k}{ h_s^L} \bm{u}_{s}^L (\bm{\bar{z}}_{s}^{L}-\bm{\bar{z}}_{s}^{L*} )^\textrm{T}
\end{eqnarray}
where $\bm{\bar{z}}_{s}^L = (\mathbf{\Theta}_{s-1}^L)^\textrm{T}\bm{\bar{x}}_{s}^L$ is
the average vector of $\textbf{Z}_{s}^{L} = \textbf{X}_{s}^{L}\mathbf{\Theta}_{s-1}^{L}$.
From (22), (40) can be further rewritten as
\begin{eqnarray}
\mathbf{\Theta}_s^L \approx \mathbf{\Theta}_{s-1}^L - \frac{k}{h_s^L} \bm{u}_{s}^L ({\mathbf{\bar{\nabla}}}_{\textbf{Z}_{s}^L})^\textrm{T}
\end{eqnarray}
where ${\mathbf{\bar{\nabla}}}_{\textbf{Z}_{s}^L} = \bm{\bar{z}}_{s}^L - \bm{\bar{z}}_{s}^{L*}$ is the average vector of $M{\mathbf{\nabla}}_{\textbf{Z}_{s}^{L}}$.
\subsection{Optimizing Hidden Layers}
Firstly, we try to derive the least squares solution of $\{\mathbf{\Theta}_s^l\}_{l=1}^{L-1}$.
Let $\mathbf{\nabla}_{\mathbf{\Theta}^{l}}$ denote $\partial J(\mathbf{\Theta}) / \partial \mathbf{\Theta}^{l}$.
By using (19) and the chain rule for $\mathbf{\Theta}^l$, we can have
\begin{eqnarray}
{\mathbf{\nabla}}_{\mathbf{\Theta}^l} =\sum_{i=1}^s (\textbf{X}_{i}^l)^\textrm{T} {\mathbf{\nabla}}_{\textbf{Z}_{i}^l}
\end{eqnarray}
where  $\mathbf{Z}_{i}^l=\textbf{X}_{i}^{l}\mathbf{\Theta}^{l}$, and
${\mathbf{\nabla}}_{\mathbf{Z}_{i}^l}$ denotes $\partial  J(\mathbf{\Theta})/\partial \textbf{Z}_{i}^l$.
By using the chain rule, ${\mathbf{\nabla}}_{\mathbf{Z}_{i}^l}$ is defined as
\begin{eqnarray}
{\mathbf{\nabla}}_{\textbf{Z}_{i}^l} = ({\mathbf{\nabla}}_{\textbf{Z}_{i}^{l+1}}(\mathbf{\Theta}^{l+1})^\textrm{T})\circ f'_l(\textbf{Z}_{i}^l)
\end{eqnarray}
Let ${\mathbf{\nabla}}_{\mathbf{\Theta}^l}=\textbf{0}$ such that
\begin{eqnarray}
\sum_{i=1}^s (\textbf{X}_{i}^l)^\textrm{T} {\mathbf{\nabla}}_{\textbf{Z}_{i}^l} =\textbf{0}
\end{eqnarray}
Unfortunately, we cannot obtain the least squares solution of $\mathbf{\Theta}_s^l$ by plugging  (43) into  (44),
since (43) is too complex and ${f'_l}(\textbf{Z}_{i}^{l})$ is generally nonlinear.

In order to tackle this problem, we present an equivalent-gradient method to redefine ${\mathbf{\nabla}}_{\textbf{Z}_{i}^l}$.
Suppose that the $l^{\textrm{th}}$ layer has the following linear least squares loss function
\begin{eqnarray}
\hat{J}_l(\mathbf{\Theta}^l)=\frac{1}{2M}\sum_{i=1}^s \lambda^{s-i} \left\|\textbf{Z}_{i}^{l}-\textbf{Z}_{i}^{l*}\right\|_\textsc{F}^2
\end{eqnarray}
where $\mathbf{Z}_{i}^{l*} \in \mathbb{R}^{M\times N_{l}} $ is the desired linear output matrix corresponding to $\mathbf{Z}_{i}^{l}$.
Obviously,  $\hat{J}_l(\mathbf{\Theta}^l)\rightarrow 0$  if $J(\mathbf{\Theta})\rightarrow 0$.
Let ${\mathbf{\hat{\nabla}}}_{\mathbf{\Theta}^l}=\partial \hat{J}_l(\mathbf{\Theta}^l)/ \partial \mathbf{\Theta}^{l}=\textbf{0}$ such that
\begin{eqnarray}
\sum_{i=1}^s  \frac{\lambda^{s-i}}{M}(\textbf{X}_{i}^{l})^\textrm{T} (\textbf{Z}_{i}^{l}-\mathbf{Z}_{i}^{l*})=\textbf{0}
\end{eqnarray}
By comparing  (44) with  (46), we can obtain an equivalent form of ${\mathbf{\nabla}}_{\textbf{Z}_{i}^l}$, which is defined as
\begin{eqnarray}
{\mathbf{\nabla}}_{\textbf{Z}_{i}^l}=\frac{ \lambda^{s-i}}{\eta^l M}(\textbf{Z}_{i}^{l}-\textbf{Z}_{i}^{l*})
\end{eqnarray}
where $\eta^l > 0$ is the gradient scaling factor. In theory,  $\eta^l$ is not fixed for different minibatches.
Considering that minibatches are generally drawn from the training dataset randomly,
we assume it is a constant in a layer.

Now, by plugging  (47) into  (44), we can easily obtain the least squares solution of $\mathbf{\Theta}_s^{l}$ as
\begin{eqnarray}
\mathbf{\Theta}_{s}^{l}=(\textbf{A}_{s}^{l})^{-1}\textbf{B}_{s}^{l}
\end{eqnarray}
where $\textbf{A}_s^{l}$  and $\textbf{B}_s^{l}$ are defined as follows
\begin{eqnarray}
\textbf{A}_s^l = \frac{1}{M} \sum_{i=1}^s  \lambda^{s-i}(\textbf{X}_{i}^l)^\textrm{T} \textbf{X}_{i}^l \\
\textbf{B}_s^l =  \frac{1}{M} \sum_{i=1}^s  \lambda^{s-i}(\textbf{X}_{i}^l)^\textrm{T} \textbf{Z}_{i}^{l*}
\end{eqnarray}

The rest of the derivation is the same as what we do in the above subsection.
Let $\textbf{P}_s^l=(\textbf{A}_s^l)^{-1}$ and define
\begin{eqnarray}
{\bm{\bar{x}}}_{s}^l=\frac{1}{M} \sum_{m=1}^M (\textbf{X}_{s(m,:)}^l)^{\textrm{T}} \\
{\bm{\bar{z}}}_{s}^{l*}=\frac{1}{M} \sum_{m=1}^M (\textbf{Z}_{s(m,:)}^{l*})^{\textrm{T}}
\end{eqnarray}
Finally, we can obtain
\begin{eqnarray}
\textbf{P}_s^l \approx \frac{1}{\lambda} \textbf{P}_{s-1}^l -  \frac{k}{\lambda h_s^l} \bm{u}_{s}^l (\bm{u}_{s}^l)^{\textrm{T}}
\end{eqnarray}
where $\bm{u}_s^l$ and $h_s^l$ are defined as
\begin{eqnarray}
\bm{u}_s^l=\textbf{P}_{s-1}^l \bm{\bar{x}}_{s}^l ~~~~ \\
h_{s}^l=\lambda+k(\bm{\bar{x}}_{s}^l)^{\textrm{T}}\bm{u}_s^l
\end{eqnarray}
The update formula of $\mathbf{\Theta}_s^l$ can be derived as
\begin{eqnarray}
\mathbf{\Theta}_s^l \approx \mathbf{\Theta}_{s-1}^l - \frac{k}{h_s^l} \bm{u}_{s}^l ({\mathbf{\bar{\nabla}}}_{\textbf{Z}_{s}^l})^\textrm{T}
\end{eqnarray}
where ${\mathbf{\bar{\nabla}}}_{\textbf{Z}_{s}^l}$ is the average vector of ${\eta^l M}{\mathbf{\nabla}}_{\textbf{Z}_{s}^{l}}$. Note that
${\mathbf{\nabla}}_{\textbf{Z}_{s}^l}$ is calculated by (43) rather than by (47), since (43) and (47) are equivalent. That means we don't have to
know $\mathbf{Z}_{s}^{l*}$.

\subsection{Practical Implementation}
In practice, we often use Pytorch or Tensorflow to build DNNs. By using their automatic differentiation (AD) package, we can easily obtain the average gradients of the minibatch loss function w.r.t. network parameters, and can efficiently implement many optimization algorithms such as SGD and Adam.
However, (41) and (56) require $\{{\mathbf{\bar\nabla}}_{\textbf{Z}_{s}^l}\}_{i=1}^{L}$, which is inconvenient to use the AD package.
In order to make our proposed algorithm easy to implement, we try to convert
(41) and (56) into a  form of SGD in this subsection.

Since $\{{\mathbf{\nabla}}_{\textbf{Z}_{s}^l}\}_{i=1}^{L}$ is only dependent on the $s^{\textrm{th}}$  minibatch,
we define an MSE loss function
\begin{eqnarray}
\tilde{J}(\mathbf{\Theta}_{s-1})=\frac{1}{2M} \left\|\textbf{Z}_{s}^{L}-\textbf{Z}_{s}^{L*}\right\|_\textsc{F}^2
\end{eqnarray}
where $\mathbf{\Theta}_{s-1}$ denotes all augmented parameter matrices of the FNN at step $s$.
Obviously, we can use it to replace (19) for calculating  ${\mathbf{\nabla}}_{\textbf{Z}_{s}^l}$, since
${\mathbf{\nabla}}_{\textbf{Z}_{s}^l}=\partial J(\mathbf{\Theta}_{s-1})/ \partial {\textbf{Z}_{s}^l} =\partial \tilde{J}(\mathbf{\Theta}_{s-1})/ \partial {\textbf{Z}_{s}^l}$.
Using the chain rule for $\mathbf{\Theta}_{s-1}$, we have
\begin{eqnarray}
\tilde{\mathbf{\nabla}}_{\mathbf{\Theta}^l_{s-1}}
= \frac{\partial \tilde{J}(\mathbf{\Theta}_{s-1})}{\partial \mathbf{\Theta}^{l}_{s-1}}
= (\textbf{X}_s^l)^\textrm{T} {\mathbf{\nabla}}_{\textbf{Z}_{s}^l}
\end{eqnarray}
Plugging (38) into (41) yields
\begin{eqnarray}
\mathbf{\Theta}_s^L \approx \mathbf{\Theta}_{s-1}^L - \frac{k}{h_s^L} \textbf{P}_{s-1}^l \bm{\bar{x}}_{s}^L ({\mathbf{\bar{\nabla}}}_{\textbf{Z}_{s}^L})^\textrm{T}
\end{eqnarray}
Since ${\mathbf{\bar{\nabla}}}_{\textbf{Z}_{s}^L}=\bm{\bar{z}}_{s}^L-\bm{\bar{z}}_{s}^{L*}$  and  $\bm{\bar{z}}_{s}^L = (\mathbf{\Theta}_{s-1}^L)^\textrm{T}\bm{\bar{x}}_{s}^L$, the above equation can be rewritten as
\begin{eqnarray}
\mathbf{\Theta}_s^L \approx \mathbf{\Theta}_{s-1}^L - \frac{k}{h_s^L}
\textbf{P}_{s-1}^L\left(\bm{\bar{x}}_{s}^L(\bm{\bar{x}}_{s}^L)^\textrm{T}\mathbf{\Theta}_{s-1}^L- \bm{\bar{x}}_{s}^L(\bm{\bar{z}}_{s}^{L*})^\textrm{T}\right)
\end{eqnarray}
From (33), (34) and $\textbf{Z}_{s}^L = \textbf{X}_s^L\mathbf{\Theta}_{s-1}^L$, (60) can be rewritten as
\begin{eqnarray}
\mathbf{\Theta}_s^L \approx \mathbf{\Theta}_{s-1}^L - \frac{1}{M h_s^L}
\textbf{P}_{s-1}^L(\textbf{X}_s^L)^\textrm{T}(\textbf{Z}_s^L - \textbf{Z}_{s}^{L*})
\end{eqnarray}
From (22), we have ${\mathbf{\nabla}}_{\textbf{Z}_{s}^L}=\frac{1}{M}(\textbf{Z}_{s}^{L}-\textbf{Z}_{s}^{L*})$. Then, by using (58), we can
rewrite (61) as \begin{eqnarray}
\mathbf{\Theta}_s^L \approx \mathbf{\Theta}_{s-1}^L -  \frac{1}{h_s^L} \textbf{P}_{s-1}^L \tilde{\mathbf{\nabla}}_{\mathbf{\Theta}^L_{s-1}}
\end{eqnarray}
Similarly, (56) can be rewritten as
\begin{eqnarray}
\mathbf{\Theta}_s^l \approx \mathbf{\Theta}_{s-1}^l -  \frac{\eta^l}{h_s^l} \textbf{P}_{s-1}^l \tilde{\mathbf{\nabla}}_{\mathbf{\Theta}^l_{s-1}}
\end{eqnarray}
Finally, by defining $\eta^L=1$, the recursive update formula of $\{\mathbf{\Theta}_{s}^{l}\}_{l=1}^{L}$ can  be unitedly written as (63).
From (63), the RLS optimization for FNNs can be viewed as a special SGD algorithm with the adaptive learning rate $\frac{\eta^l}{ h_s^l} \textbf{P}_{s-1}^l$.

In conclusion, an FNN with the RLS optimization can be summarized in Algorithm 1.
Note here that the loss function won't have to be (57) if the RLS optimization is only used for training hidden layers.
In other words, our algorithms can be also used in combination with other first-order algorithms. For example, we can
use cross-entropy as the loss function, use Adam for training the output layer, and use  the RLS optimization for training
hidden layers. This is because $\textbf{P}_{s-1}^l$ and $h_s^l$ only depend on the inputs of the $l^{\textrm{th}}$ layer.
In addition, from (42) and (43), $\tilde{\mathbf{\nabla}}_{\mathbf{\Theta}^l_{s-1}}$ is generally derived from
${\mathbf{\nabla}}_{\textbf{Z}_{i}^{l+1}}$ by using the backward propagation method.

\SetKw{KwInput}{Input}
\SetKw{KwInitial}{Initialize}
\begin{algorithm}
\caption{FNN with the RLS Optimization}
\KwInput: {$\lambda$, $k$, $\{\eta^l\}_{l=1}^{L}$}\;
\KwInitial: $\{\mathbf{\Theta}^l_0, \textbf{P}_0^l\}_{l=1}^{L}$, $s=1$\;
\LinesNumbered
\While{stopping criterion not met}{
   Get the minibatch $(\textbf{Y}_s^0, \textbf{Y}^*_s)$ from the training dataset\;
   \For{$l=1~to~L$}{
      Compute $\textbf{Y}_s^l$ by (2)\;
   }
   Compute $\tilde{J}(\mathbf{\Theta}_{s-1})$\ by  (57)\;
   \For{$l=L~to~1$}{
      Compute ${\tilde{\mathbf{\nabla}}}_{\mathbf{\Theta}_{s-1}^{l}}$ by the AD package\;
      Compute  $\bm{\bar{x}}^l_s$, $\bm{u}^l_s$, $h^l_s$ by  (51), (54), (55)\;
      Update $\mathbf{\Theta}^l_{s-1}$, $\textbf{P}^l_{s-1}$ by  (63), (53)\;
   }
   $s= s+1$;
}
\end{algorithm}

\section{RLS Optimization for Other DNNs}
In this section, we first introduce the derivation of the RLS optimization for CONV layers in CNNs. Then, we introduce the derivation
of the RLS optimization for output and hidden layers in RNNs, and present the detailed pseudocode. We also introduce the RLS optimization for LSTMs
briefly.

\subsection{Optimizing CNNs}
As reviewed in Section \uppercase\expandafter{\romannumeral2}.\textit{B}, the latter part of a CNN generally consists of
some FC layers. Therefore, the least squares loss function and the RLS optimization for FC layers in CNNs are the same as those in FNNs.
Here, we only consider the RLS optimization for CONV layers. The derivation is similar to the derivation presented in
Sections \uppercase\expandafter{\romannumeral3}.\textit{C} and \uppercase\expandafter{\romannumeral3}.\textit{D}.

Firstly, based on (19) and (6), we use the chain rule for  ${\mathbf{\nabla}}_{\mathbf{\Theta}^l}=\partial J(\mathbf{\Theta})/ \partial \mathbf{\Theta}^{l}$.
Let ${\mathbf{\nabla}}_{\mathbf{\Theta}^l}=\textbf{0}$ such that
\begin{eqnarray}
\sum_{i=1}^s\sum_{u=1}^{U_l} \sum_{v=1}^{V_l} (\textbf{X}_{i(:,:,u,v)}^l)^\textrm{T} {\mathbf{\nabla}}_{\textbf{Z}_{i(:,:,u,v)}^l} = \textbf{0}
\end{eqnarray}
where ${\mathbf{\nabla}}_{\mathbf{Z}_{i(:,:,u,v)}^l}$ denotes $\partial J(\mathbf{\Theta})/\partial\textbf{Z}_{i(:,:,u,v)}^l$.

Secondly, we suppose that the current CONV layer has the following least squares loss function
\begin{eqnarray}
\hat{J}_l(\mathbf{\Theta}^l)=\frac{1}{2MU_lV_l}\sum_{i=1}^s \lambda^{s-i} \left\|\textbf{Z}_{i}^{l*}-\textbf{Z}_{i}^{l}\right\|_\textsc{F}^2
\end{eqnarray}
where $\mathbf{Z}_{i}^{l*} \in \mathbb{R}^{M\times C_{l}\times U_{l}\times V_{l}}$ is the linear desired output matrix
of the $i^{\textrm{th}}$ minibatch in this layer.
Let ${\mathbf{\hat{\nabla}}}_{\mathbf{\Theta}^l} =\partial \hat{J}_l(\mathbf{\Theta}^l)/ \partial \mathbf{\Theta}^{l}=\textbf{0}$ such that
\begin{eqnarray}
\sum_{i=1}^s\sum_{u=1}^{U_l} \sum_{v=1}^{V_l} (\textbf{X}_{i(:,:,u,v)}^l)^\textrm{T} {\hat{\mathbf{\nabla}}}_{\textbf{Z}_{i(:,:,u,v)}^l} = \textbf{0}
\end{eqnarray}
where ${\hat{\mathbf{\nabla}}}_{\mathbf{Z}_{i(:,:,u,v)}^l}=\partial \hat{J}_l(\mathbf{\Theta})/\partial\textbf{Z}_{i(:,:,u,v)}^l$ is defined as
\begin{eqnarray}
{\hat{\mathbf{\nabla}}}_{\mathbf{Z}_{i(:,:,u,v)}^l} =
- \frac{\lambda^{s-i}}{MU_lV_l}(\textbf{Z}_{i(:,:,u,v)}^{l*}-\mathbf{Z}_{i(:,:,u,v)}^{l})
\end{eqnarray}
Then, by comparing (64) with (66), we can define the equivalent gradient of  ${\mathbf{\nabla}}_{\mathbf{Z}_{i(:,:,u,v)}^l}$ as
\begin{eqnarray}
{\mathbf{\nabla}}_{\mathbf{Z}_{i(:,:,u,v)}^l} =
- \frac{\lambda^{s-i}}{\eta^l MU_lV_l}(\textbf{Z}_{i(:,:,u,v)}^{l*}-\mathbf{Z}_{i(:,:,u,v)}^{l})
\end{eqnarray}

Thirdly, by plugging (68) into (64), we can obtain the least squares solution of $\mathbf{\Theta}_s^{l}$, which is written as
\begin{eqnarray}
\mathbf{\Theta}_{s}^{l}=(\textbf{A}_{s}^{l})^{-1}\textbf{B}_{s}^{l}
\end{eqnarray}
where $\textbf{A}_s^{l}$  and $\textbf{B}_s^{l}$ are defined as follows
\begin{eqnarray}
\textbf{A}_s^l = \frac{1}{MU_lV_l} \sum_{i=1}^s \sum_{u=1}^{U_l} \sum_{v=1}^{V_l}\lambda^{s-i}(\textbf{X}_{i(:,:,u,v)}^l)^\textrm{T} \textbf{X}_{i(:,:,u,v)}^l  \\
\textbf{B}_s^l =  \frac{1}{MU_lV_l} \sum_{i=1}^s \sum_{u=1}^{U_l} \sum_{v=1}^{V_l}\lambda^{s-i}(\textbf{X}_{i(:,:,u,v)}^l)^\textrm{T} \textbf{Z}_{i(:,:,u,v)}^{l*}
\end{eqnarray}

The rest of the derivation is similar to the recursive derivation of (49) and (48). Let $\textbf{P}_s^l=(\textbf{A}_s^l)^{-1}$, and define
\begin{eqnarray}
{\bm{\bar{x}}}_{s}^l=\frac{1}{M U_l V_l} \sum_{m=1}^M \sum_{u=1}^{U_l} \sum_{v=1}^{V_l} (\textbf{X}_{s(m,:,u,v)}^l)^{\textrm{T}}  \\
{\bm{\bar{z}}}_{s}^{l*}=\frac{1}{M U_l V_l} \sum_{m=1}^M \sum_{u=1}^{U_l} \sum_{v=1}^{V_l} (\textbf{Z}_{s(m,:,u,v)}^{l*})^{\textrm{T}}
\end{eqnarray}
Then, similar to (25) and (26), (70) and (71) can be approximately rewritten as follows
\begin{eqnarray}
\textbf{A}_{s}^l \approx \lambda\textbf{A}_{s-1}^l + k{\bm{\bar{x}}}_{s}^l(\bm{\bar{x}}_{s}^l)^\textrm{T} \\
\textbf{B}_{s}^l \approx \lambda\textbf{B}_{s-1}^l + k{\bm{\bar{x}}}_{s}^l(\bm{\bar{z}}^{l*}_{s})^\textrm{T}
\end{eqnarray}
Finally, we can obtain $\textbf{P}_s^l$, $\bm{u}^l_s$, $h_s^l$  and $\mathbf{\Theta}_s^l $, which have the same forms
as (53), (54), (55) and (63), respectively.

The pseudocode of a CNN with the RLS optimization is similar to Algorithm 1. For brevity, we omit it here.

\subsection{Optimizing RNNs}
As reviewed in Section \uppercase\expandafter{\romannumeral2}.\textit{C}, RNNs generally process sequential data, and use the BPTT method to update network parameters.
Therefore, the RLS optimization for RNNs is a little bit different from those for FNNs and CNNs.

Firstly, considering the minibatch training dataset $\{(\{\textbf{Y}_{i,t}^{0}\}_{t=1}^{T},\{\textbf{Y}_{i,t}^{*}\}_{t=t_0}^{T})\}_{i=1}^{s}$, we define the linear least squares loss function as
\begin{eqnarray}
J(\mathbf{\Theta})=\frac{1}{2M} \sum_{i=1}^s\sum_{t=t_0}^T \lambda^{s-i} \left\|\textbf{Z}_{i,t}^{L}- \textbf{Z}_{i,t}^{L*}\right\|_\textsc{F}^2
\end{eqnarray}
where $\mathbf{\Theta}$ denotes all augmented parameter matrices of the RNN, $\mathbf{Z}_{i,t}^{L*}=f_L^{-1}(\textbf{Y}_{i,t}^*)$ and $\mathbf{Z}_{i,t}^{L} = \textbf{X}_{i,t}^{L}\mathbf{\Theta}^{L}$.

Secondly, we try to use (74) to derive the RLS solution of $\mathbf{\Theta}_s^{L}$. By using the method used in
Section \uppercase\expandafter{\romannumeral3}.\textit{B}, we can get
\begin{eqnarray}
\mathbf{\Theta}_s^L =(\textbf{A}_s^L)^{-1} \textbf{B}_s^L
\end{eqnarray}
where $\textbf{A}_s^L$ and $\textbf{B}_s^L$ are defined as follows
\begin{eqnarray}
\textbf{A}_s^L = \frac{1}{M} \sum_{i=1}^s\sum_{t=t_0}^T  \lambda^{s-i}(\textbf{X}_{i,t}^L)^\textrm{T} \textbf{X}_{i,t}^L  \\
\textbf{B}_s^L =  \frac{1}{M} \sum_{i=1}^s\sum_{t=t_0}^T  \lambda^{s-i}(\textbf{X}_{i,t}^L)^\textrm{T} \textbf{Z}_{i,t}^{L*}
\end{eqnarray}
Let $\textbf{P}_s^L=(\textbf{A}_s^L)^{-1}$, and define
\begin{eqnarray}
{\bm{\bar{x}}}_{s}^L=\frac{1}{MT_L} \sum_{t=t_0}^T \sum_{m=1}^M (\textbf{X}_{s,t(m,:)}^L)^{\textrm{T}} \\
{\bm{\bar{z}}}_{s}^{L*}=\frac{1}{MT_L} \sum_{t=t_0}^T \sum_{m=1}^M (\textbf{Z}_{s,t(m,:)}^{L*})^{\textrm{T}}
\end{eqnarray}
where $T_L=\tau-t_0+1$. Then, similar to (29) and (30), (76) and (77) can be approximately rewritten as follows
\begin{eqnarray}
\textbf{A}_{s}^L \approx \lambda\textbf{A}_{s-1}^L + kT_L{\bm{\bar{x}}}_{s}^L(\bm{\bar{x}}_{s}^L)^\textrm{T} \\
\textbf{B}_{s}^L \approx \lambda\textbf{B}_{s-1}^L + kT_L{\bm{\bar{x}}}_{s}^L(\bm{\bar{z}}^{L*}_{s})^\textrm{T}
\end{eqnarray}
By using the Sherman-Morrison formula for (82), we can easily derive
\begin{eqnarray}
\textbf{P}_s^L \approx \frac{1}{\lambda} \textbf{P}_{s-1}^L -  \frac{kT_L}{\lambda h_s^L} \bm{u}_{s}^L (\bm{u}_{s}^L)^{\textrm{T}}
\end{eqnarray}
where $\bm{u}_s^L$ and $h_s^L$ are defined as
\begin{eqnarray}
\bm{u}_s^L=\textbf{P}_{s-1}^L \bm{\bar{x}}_{s}^L  ~~~~~~&\\
h_{s}^L=\lambda+ kT_L(\bm{\bar{x}}_{s}^L)^{\textrm{T}}\bm{u}_s^L
\end{eqnarray}
Let ${\tilde{\mathbf{\nabla}}}_{\mathbf{\Theta}_{s-1}^{L}}$ to be the gradient of the following loss function
\begin{eqnarray}
\tilde{J}(\mathbf{\Theta}_{s-1})=\frac{1}{2M}\sum_{t=t_0}^T \left\|\textbf{Z}_{s,t}^{L}-\textbf{Z}_{s,t}^{L*}\right\|_\textsc{F}^2
\end{eqnarray}
Derived by using the same techniques in Sections \uppercase\expandafter{\romannumeral3}.\textit{B} and \uppercase\expandafter{\romannumeral3}.\textit{D},
$\mathbf{\Theta}_s^L$ has the same recursive form as (62).

Thirdly, using the methods in Sections \uppercase\expandafter{\romannumeral3}.\textit{C} and \uppercase\expandafter{\romannumeral3}.\textit{D}, we try to
derive the RLS solution of $\{(\mathbf{\Theta}_s^{l_w}, \mathbf{\Theta}_s^{l_v})\}_{l=1}^{L-1}$. First of all, we suppose the
$l^{th}$ layer has the loss function
\begin{equation}
\begin{aligned}
\hat{J}_l(\mathbf{\Theta}^l)= \frac{1}{2M}\sum_{i=1}^s\sum_{t=1}^T \sum_{\nu \in \Psi} \lambda^{s-i} \left\|\textbf{Z}_{i,t}^{l_\nu}-\textbf{Z}_{i,t}^{l_\nu*}\right\|_\textsc{F}^2
\end{aligned}
\end{equation}
where $\Psi=\{w,v\}$, and $\mathbf{Z}_{i,t}^{l_\nu*} \in \mathbb{R}^{M\times N_{l}}$ is
the desired linear output matrix corresponding  to $\mathbf{Z}_{i,t}^{l_\nu}$.
Then, using the equivalent gradient of $\partial J(\mathbf{\Theta})/\partial \textbf{Z}_{i,t}^{l_\nu}$ derived by (88), we can
obtain
\begin{eqnarray}
\mathbf{\Theta}_{s}^{l_\nu}=(\textbf{A}_{s}^{l_\nu})^{-1}\textbf{B}_{s}^{l_\nu}
\end{eqnarray}
where $\textbf{A}_s^{l_\nu}$  and $\textbf{B}_s^{l_\nu}$ are defined as follows
\begin{eqnarray}
\textbf{A}_s^{l_\nu} = \frac{1}{M} \sum_{i=1}^s\sum_{t=1}^T  \lambda^{s-i}(\textbf{X}_{i,t}^{l_\nu})^\textrm{T} \textbf{X}_{i,t}^{l_\nu}  \\
\textbf{B}_s^{l_\nu} =  \frac{1}{M} \sum_{i=1}^s\sum_{t=1}^T  \lambda^{s-i}(\textbf{X}_{i,t}^{l_\nu})^\textrm{T} \textbf{Z}_{i,t}^{{l_\nu}*}
\end{eqnarray}
Let $\textbf{P}_s^{l_\nu}=(\textbf{A}_s^{l_\nu})^{-1}$, and define
\begin{eqnarray}
{\bm{\bar{x}}}_{s}^{l_\nu}=\frac{1}{MT_l} \sum_{t=1}^T \sum_{m=1}^M (\textbf{X}_{s,t(m,:)}^{l_\nu})^{\textrm{T}}  \\
{\bm{\bar{z}}}_{s}^{{l_\nu}*}=\frac{1}{MT_l} \sum_{t=1}^T \sum_{m=1}^M (\textbf{Z}_{s,t(m,:)}^{{l_\nu}*})^{\textrm{T}}
\end{eqnarray}
where $T_l=T$.
The rest of the derivation is similar to the derivation of $\textbf{P}_s^L$ and $\mathbf{\Theta}_s^L$. We can derive
$\textbf{P}_s^{l_\nu}$  as
\begin{eqnarray}
\textbf{P}_s^{l_\nu} \approx \frac{1}{\lambda} \textbf{P}_{s-1}^{l_\nu} -  \frac{kT_l}{\lambda h_s^{l_\nu}} \bm{u}_{s}^{l_\nu} (\bm{u}_{s}^{l_\nu})^{\textrm{T}}
\end{eqnarray}
where $\bm{u}_s^{l_\nu}$ and $h_s^{l_\nu}$ are defined as
\begin{eqnarray}
\bm{u}_s^{l_\nu}=\textbf{P}_{s-1}^{l_\nu} \bm{\bar{x}}_{s}^{l_\nu}  ~~~~~&\\
h_{s}^{l_\nu}=\lambda+ kT_{l}(\bm{\bar{x}}_{s}^{l_\nu})^{\textrm{T}}\bm{u}_s^{l_\nu}
\end{eqnarray}
$\mathbf{\Theta}_s^{l_\nu}$ is finally derived as
\begin{eqnarray}
\mathbf{\Theta}_s^{l_\nu} \approx \mathbf{\Theta}_{s-1}^{l_\nu} -  \frac{\eta^{l_\nu}}{h_s^{l_\nu}} \textbf{P}_{s-1}^{l_\nu} {\tilde{\mathbf{\nabla}}}_{\mathbf{\Theta}_{s-1}^{l_\nu}}
\end{eqnarray}
where $\eta^{l_\nu} > 0$ is the gradient scaling factor.

In conclusion, a stacked RNN with the RLS optimization can be summarized in Algorithm 2.
As reviewed in Section \uppercase\expandafter{\romannumeral2}.\textit{C},  LSTMs can be described by using
$\mathbf{\Theta}_{s-1}^L$ and $\{(\mathbf{\Theta}_{s-1}^{l_w}, \mathbf{\Theta}_{s-1}^{l_v})\}_{l=1}^{L-1}$ as well.
Thus,  by replacing ``(8)" in line 9 with ``(18)", Algorithm 2 can be also used for optimizing stacked LSTMs.
For brevity, we don't repeat the derivation of the RLS optimization for LSTMs.

\SetKw{KwInput}{Input}
\SetKw{KwInitial}{Initialize}
\begin{algorithm}
\caption{RNN with the RLS Optimization}
\KwInput: {$\lambda$, $k$, $\{(\eta^{l_w}, \eta^{l_v})\}_{l=1}^{L-1}$}\;
\KwInitial: $\{(\mathbf{\Theta}_0^{l_w}, \mathbf{\Theta}_0^{l_v},\textbf{P}_0^{l_w}, \textbf{P}_0^{l_v})\}_{l=1}^{L-1}$, $\mathbf{\Theta}^L_0$,  $\textbf{P}^L_0$, $s=1$\;
\LinesNumbered
\While{stopping criterion not met}{
   Get the minibatch $(\{\textbf{Y}_{s,t}^0\}_{t=1}^T, \{\textbf{Y}^*_{s,t}\}_{t=t_0}^T)$ from
   the sequential training dataset and set $\{\textbf{Y}_{s,0}^l=\textbf{0}\}_{l=1}^{L-1}$\;
   \For{$l=1~to~L-1$}{
      \For{$t=1~to~T$}{
         Compute $\textbf{Y}_{s,t}^l$ by (7)\;
      }
   }
   \For{$t=t_0~to~T$}{
         Compute $\textbf{Y}_{s,t}^l$ by (8)\;
   }
   Compute $\tilde{J}(\mathbf{\Theta}_{s-1})$ by (88)\;
   Compute ${\tilde{\mathbf{\nabla}}}_{\mathbf{\Theta}_{s-1}^{L}}$ by the AD package\;
   Compute $\bm{\bar{x}}^L_s$, $\bm{u}^L_s$, $h^L_s$ by (80), (85), (86)\;
   Update $\mathbf{\Theta}^L_{s-1}$, $\textbf{P}^L_{s-1}$ by (62), (84)\;
   \For{$l=L-1~to~1$}{
      \For{$\nu$ in  $\{w,v\}$}{
      Compute ${\tilde{\mathbf{\nabla}}}_{\mathbf{\Theta}_{s-1}^{l_{\nu}}}$ by the AD package\;
      Compute $\bm{\bar{x}}^{l_{\nu}}_s$, $\bm{u}^{l_{\nu}}_s$, $h^{l_{\nu}}_s$ by (92), (95), (96)\;
      Update $\mathbf{\Theta}^{l_{\nu}}_{s-1}$, $\textbf{P}^{l_{\nu}}_{s-1}$ by  (97), (94)\;
      }
   }
   $s= s+1$;
}
\end{algorithm}

\section{Analysis and Improvement}
In this section, we first analyze the time and space complexity of our proposed algorithms. Then, we discuss
some possible improvements on our proposed algorithms by using existing studies in DL and RLS research
communities, and present two improved methods for our algorithms.

\subsection{Complexity Analysis}
As mentioned in Section \uppercase\expandafter{\romannumeral1}, previous RLS optimization algorithms have fast
convergence, but have high time and space complexities and many additional preconditions. By using the average-approximation
method and the  equivalent-gradient method, our algorithms can be viewed as a special SGD algorithm, which use
$\frac{\eta^l}{ h_s^l} \textbf{P}_{s-1}^l$ or $\frac{\eta^{l_\nu}}{ h_s^{l_\nu}} \textbf{P}_{s-1}^{l_\nu}$ as the learning rate.
Compared to previous RLS optimization algorithms, our algorithms are straightforward, simple and easy to implement.
Although they require extra storage and calculation for $\textbf{P}_{s-1}^l$ at each iteration,
their time and space complexities are still acceptable.
The time and space complexity comparisons between them and the conventional SGD algorithm are summarized in Tables I and II,
where  $\hat{C}_{l-1}$ denotes $C_{l-1}H_lW_l$, $\hat{N}_{l-1}$ denotes $N_{l-1}+N_l$,
$\tilde{N}^2_{l-1}$ denotes $N^2_{l-1}+N^2_l$, the superscript $a$ denotes that the FC layer is used in FNNs or CNNs,
and the superscript $b$ denotes that the FC layer is used in RNNs or LSTMs.
Note here that we ignore lower order terms for brevity.

From Table I,  for the FC layer in FNNs or CNNs,  the time complexity of the
RLS optimization is about 2 to 32 times  that of the conventional SGD algorithm, since $N_{l-1}$ is 256, 512 or 1024 and $M$ is 32, 64 or 128 in general.
Apart from this, for any other type of layers, the time complexity of the RLS optimization is usually less than twice
that of the conventional SGD algorithm, since $\hat{C}_{l-1}$ is less than $MU_lV_l$ and $\hat{N}_{l-1}$ is less than $MT_l$ in general.
From Table II, for FC, CONV, RECUR and LSTM layers, if $N_{l-1}=N_l$, $C_{l-1}=C_l$ and $H_l=W_l=2$,
the space complexities of the RLS optimization will be 2, 5, 2 and 1.25 times those of the conventional SGD algorithm, respectively.
Thus, the time and space complexities of our algorithms are generally the same order as those of the conventional SGD algorithm.
\begin{table}[!htp]
\renewcommand{\arraystretch}{1.5}
\centering
\caption{Time Complexity Comparison Between RLS and SGD}
\setlength{\tabcolsep}{3pt}
\scriptsize
\begin{tabular}{|p{1.5cm}<{\centering}|p{1.7cm}<{\centering}|p{3.1cm}<{\centering}|p{1.6cm}<{\centering}|}
\hline
 & SGD & RLS & RLS/SGD \\
\hline
FC Layer $^a$  & $N_{l-1}MN_l$ & $N_{l-1}(M+N_{l-1})N_l$ & $1+ \frac{N_{l-1}}{M}$\\
\hline
CONV Layer & $\hat{C}_{l-1} MU_lV_l C_l$ & $\hat{C}_{l-1}(MU_lV_l+\hat{C}_{l-1})C_l$ & $1+ \frac{\hat{C}_{l-1}}{MU_lV_l}$\\
\hline
FC Layer $^b$  & $N_{l-1}MT_lN_l$ & $N_{l-1}(MT_l+N_{l-1})N_l$ & $1+ \frac{N_{l-1}}{MT_l}$\\
\hline
RECUR Layer & $\hat{N}_{l-1}MT_l N_l$ & $\hat{N}_{l-1}MT_lN_l+ \tilde{N}^2_{l-1}N_l$ & $1+ \frac{\tilde{N}^2_{l-1}}{\hat{N}_{l-1}MT_l}$\\
\hline
LSTM Layer & $4\hat{N}_{l-1}MT_l N_l$& $4\hat{N}_{l-1}MT_lN_l+ 4\tilde{N}^2_{l-1}N_l$ & $1+ \frac{\tilde{N}^2_{l-1}}{\hat{N}_{l-1}MT_l}$\\
\hline
\end{tabular}
\end{table}
\begin{table}[h!]
\renewcommand{\arraystretch}{1.5}
\centering
\caption{Space Complexity Comparison Between RLS and SGD}
\setlength{\tabcolsep}{3pt}
\scriptsize
\begin{tabular}{|p{1.5cm}<{\centering}|p{1.7cm}<{\centering}|p{3.1cm}<{\centering}|p{1.6cm}<{\centering}|}
\hline
& SGD & RLS & RLS/SGD \\
\hline
FC Layer & $N_{l-1}N_l$ & $N_{l-1}(N_l+N_{l-1})$ & $1+ \frac{N_{l-1}}{N_l}$\\
\hline
CONV Layer & $\hat{C}_{l-1}C_l$ & $\hat{C}_{l-1}(C_l+\hat{C}_{l-1})$ & $1+ \frac{\hat{C}_{l-1}}{C_l}$\\
\hline
RECUR Layer & $\hat{N}_{l-1}N_l$ & $\hat{N}_{l-1}N_l+\tilde{N}^2_{l-1}$ & $1+ \frac{\tilde{N}^2_{l-1}}{\hat{N}_{l-1}N_l}$\\
\hline
LSTM Layer & $4\hat{N}_{l-1}N_l$& $4\hat{N}_{l-1}N_l+\tilde{N}^2_{l-1}$ & $1+ \frac{\tilde{N}^2_{l-1}}{4\hat{N}_{l-1}N_l}$\\
\hline
\end{tabular}
\end{table}

\subsection{Performance Improvement}
Our algorithms are established on the ordinary RLS algorithm and the conventional SGD algorithm. In the last decades, there have been
many studies on improving the ordinary RLS algorithm in many ways, such as reducing its complexity \cite{Claser2016}, enhancing its robustness \cite{Sadigh2020},
preventing overfitting \cite{Eksioglu2010} and adjusting the forgetting factor adaptively \cite{Albu2012}. As mentioned in Section \uppercase\expandafter{\romannumeral1},
there have also been many studies on improving the conventional SGD algorithm.
Thus, our algorithms still have a huge room for improvement,
and utilizing these existing studies should be the most direct and effective way.
Here, we introduce two improved methods for our algorithms briefly.

The first improved method is to prevent our algorithms from overfitting by using  regularization techniques. It is well known that
DNNs are prone to overfitting. In fact, the ordinary RLS is prone to overfitting, too. In \cite{Eksioglu2011},
Eksioglu and Tanc  proposes a general convex regularization method for the ordinary RLS. By using this method,
we can modify (63) as
\begin{eqnarray}
\mathbf{\Theta}_s^l \approx \mathbf{\Theta}_{s-1}^l -  \frac{\eta^l}{ h_s^l} \textbf{P}_{s-1}^l {\tilde{\mathbf{\nabla}}}_{\mathbf{\Theta}_{s-1}^{l}} -
 \gamma\textbf{P}_{s}^l \mathrm{sgn}(\mathbf{\Theta}_{s-1}^{l})
\end{eqnarray}
where $\gamma$ is the regularization factor,
and $\mathrm{sgn}(\cdot)$ is the sign function. Similarly, (97) can be also
modified into the form similar to (98).  We omit it here for brevity.

The second improved method is to accelerate our algorithms by using the momentum method \cite{Polyak1964}.
In DL, momentum is widely used in many optimization algorithms. It can deal with the poor conditioning of the Hessian matrix
and reduce the variance of the stochastic gradient \cite{Goodfellow2016}. By using it, we can further modify (98) as follows
\begin{eqnarray}
\mathbf{\Omega}_s^l = \alpha\mathbf{\Omega}_{s-1}^l -  \frac{\eta^l}{ h_s^l} \textbf{P}_{s-1}^l {\tilde{\mathbf{\nabla}}}_{\mathbf{\Theta}_{s-1}^{l}}~~ \\
\mathbf{\Theta}_s^l \approx \mathbf{\Theta}_{s-1}^l + \mathbf{\Omega}_s^l  -
 \gamma\textbf{P}_{s}^l \mathrm{sgn}(\mathbf{\Theta}_{s-1}^{l})
\end{eqnarray}
where $\mathbf{\Omega}_s^l$ is the velocity matrix of the $l^{\textrm{th}}$ layer at step $s$, and $\alpha$ is the momentum factor.  Similarly, (97) can be also
modified into forms similar to (99) and (100).

\section{Experiments}
In this section, by using the Adam algorithm for comparison, we demonstrate the effectiveness of our algorithms for training FNNs, CNNs and LSTMs.
In addition, we also investigate the influences of hyperparameters on the convergence performances
of our algorithms experimentally.

\subsection{Experimental Settings}
We prepare three experiments in total, namely, FNNs on the MNIST dataset, CNNs on the CIFAR-10 dataset, and
LSTMs on the IMDB dataset. For brevity, the experiment on RNNs is not presented here, since LSTMs are used more
widely in practice. Here, we use the Adam algorithm for comparison, since it is perhaps the most excellent
first-order optimization algorithm for training DNNs.
In each experiment, we evaluate the convergence performances of our algorithms
together with or without Adam. We also investigate the influences of hyperparameters  $k$ and $\eta^l$ (or $\{(\eta^{l_w}, \eta^{l_v})\}_{l=1}^{L-1}$)
 on the performances of our algorithms.
We don't investigate the hyperparameter $\lambda$, since its influence on RLS-type algorithms has been investigated intensively \cite{Albu2012},
although it also has a significant influence on the performances of our algorithms.
\begin{figure*}[htbp]
\centering
\subfigure[FNNs on MNIST]{\includegraphics[width=0.32\textwidth]{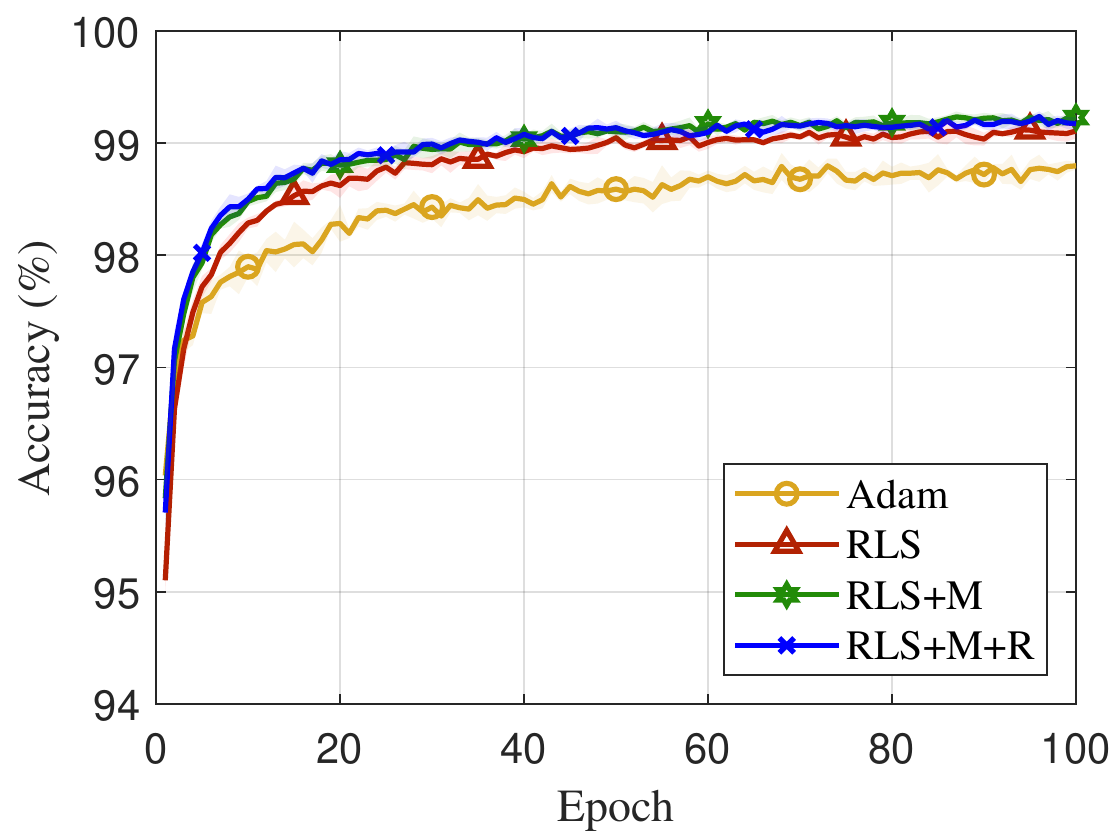}}
\subfigure[CNNs on CIFAR-10]{\includegraphics[width=0.32\textwidth]{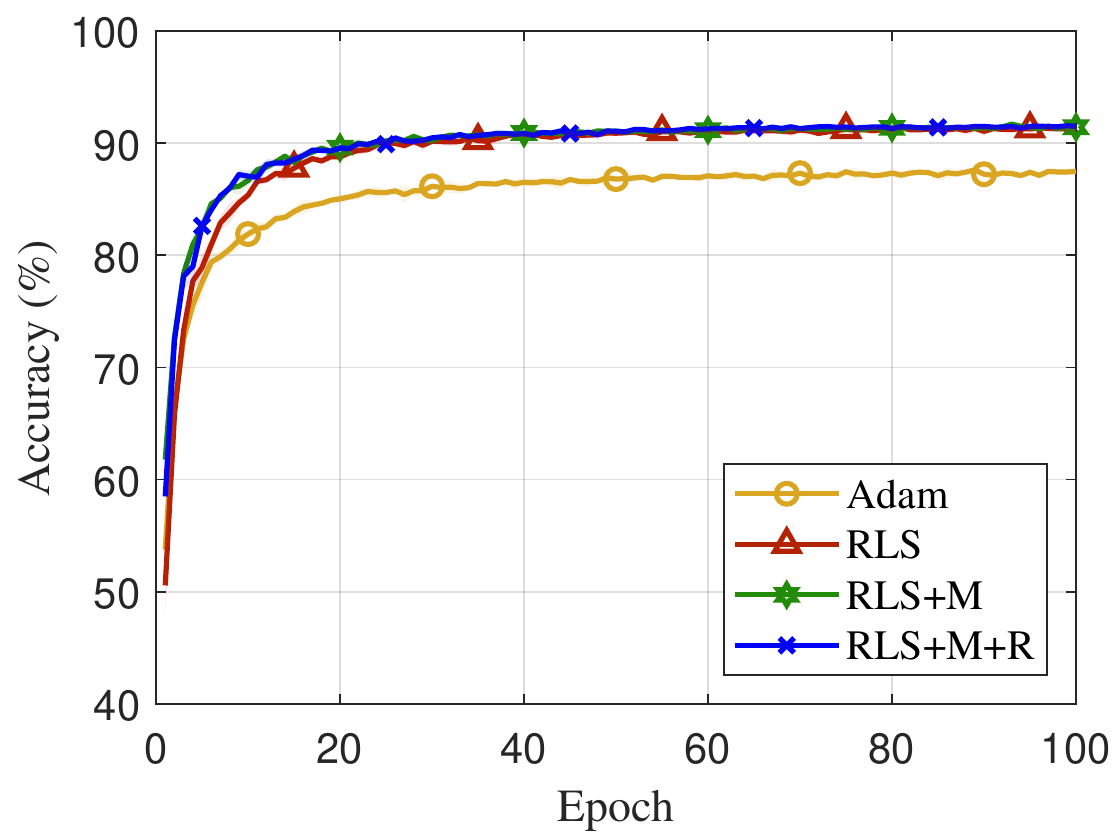}}
\subfigure[LSTMs on IMDB]{\includegraphics[width=0.32\textwidth]{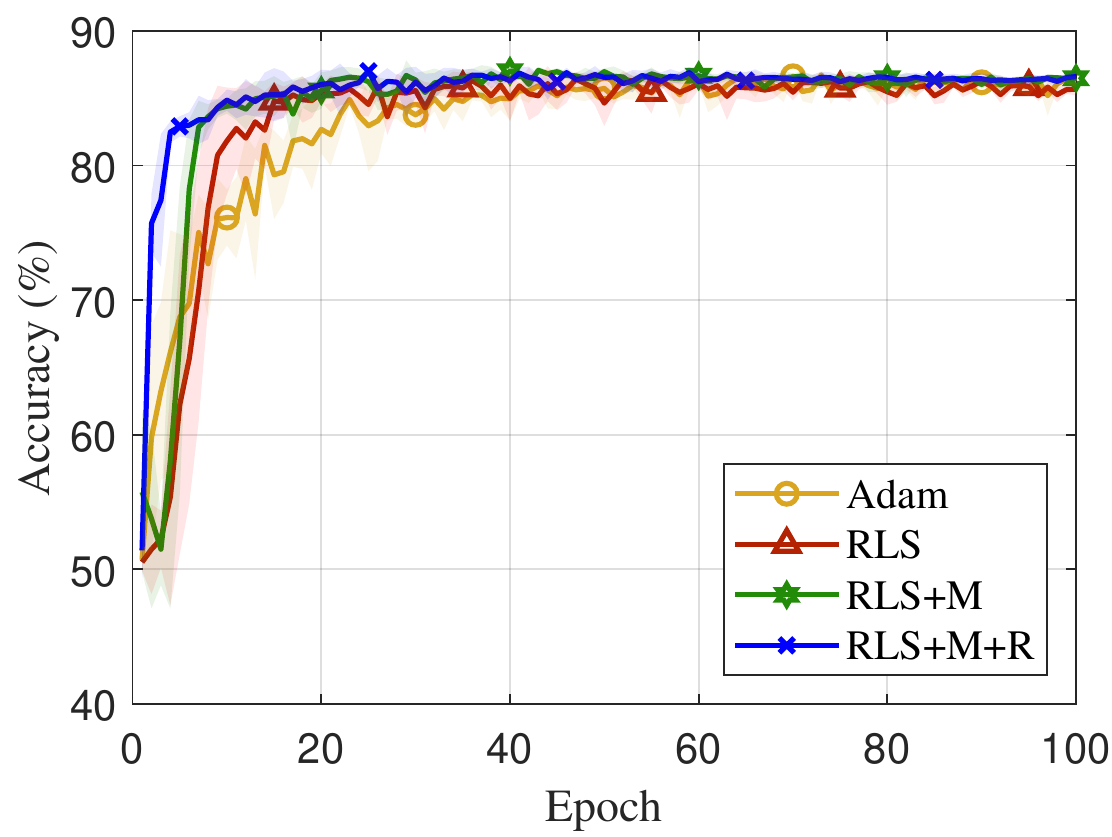}}
\caption{Performance evaluation of our three algorithms used alone against the Adam algorithm on MNIST, CIFAR-10 and IMDB with the MSE loss.}
\end{figure*}
\begin{figure*}[htbp]
\centering
\subfigure[FNNs on MNIST]{\includegraphics[width=0.32\textwidth]{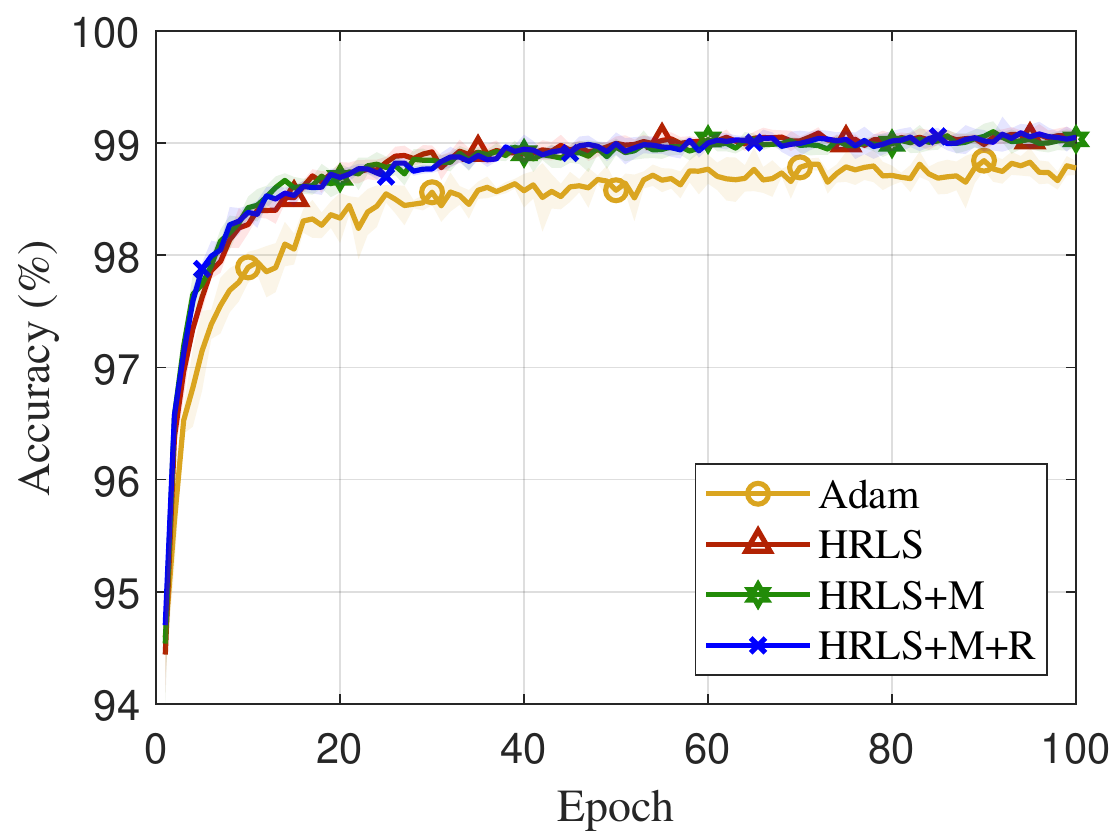}}
\subfigure[CNNs on CIFAR-10]{\includegraphics[width=0.32\textwidth]{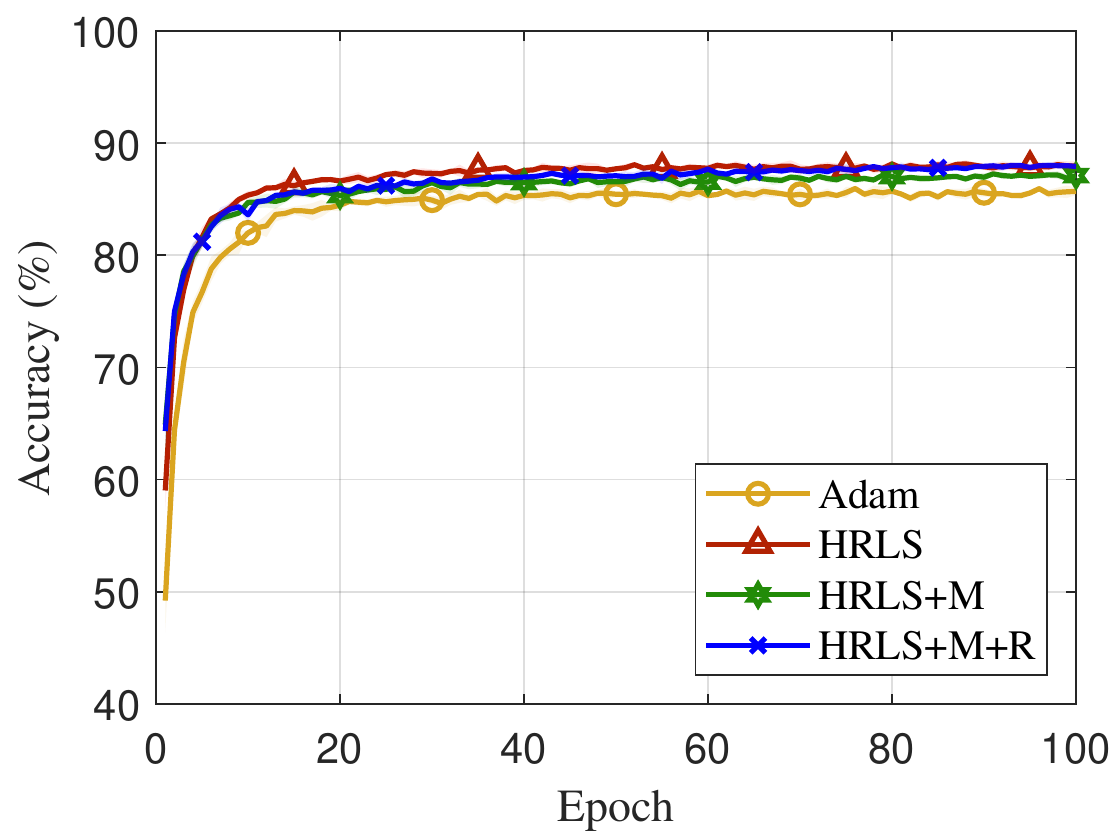}}
\subfigure[LSTMs on IMDB]{\includegraphics[width=0.32\textwidth]{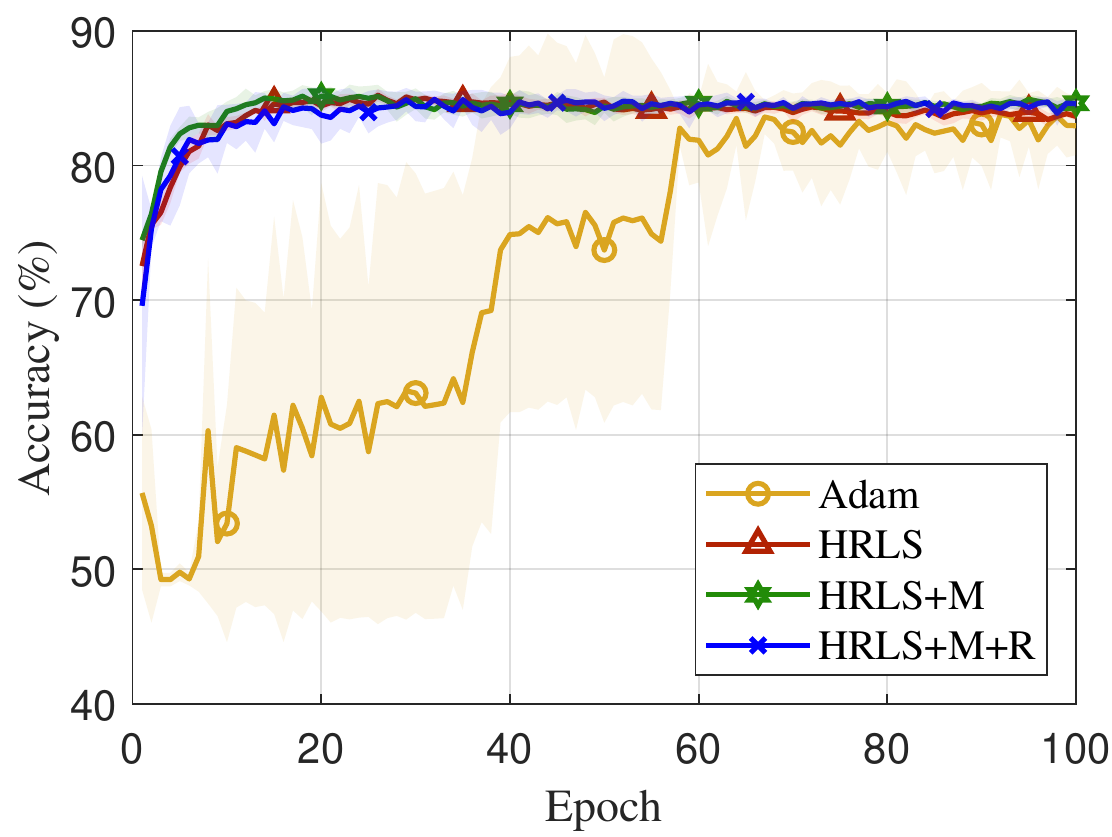}}
\caption{Performance evaluation of our algorithms used partly against the Adam algorithm on MNIST, CIFAR-10 and IMDB with the cross-entropy loss.}
\end{figure*}
\begin{figure*}[htbp]
\centering
\subfigure[FNNs on MNIST]{\includegraphics[width=0.32\textwidth]{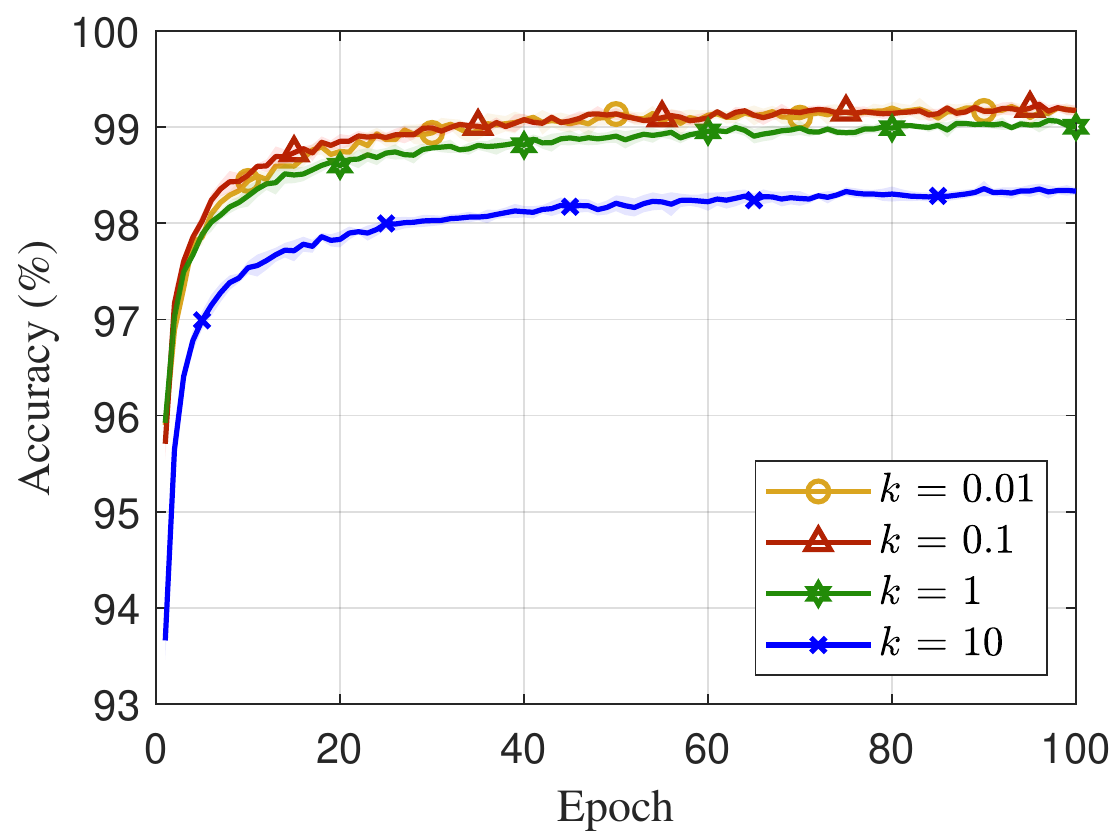}}
\subfigure[CNNs on CIFAR-10]{\includegraphics[width=0.32\textwidth]{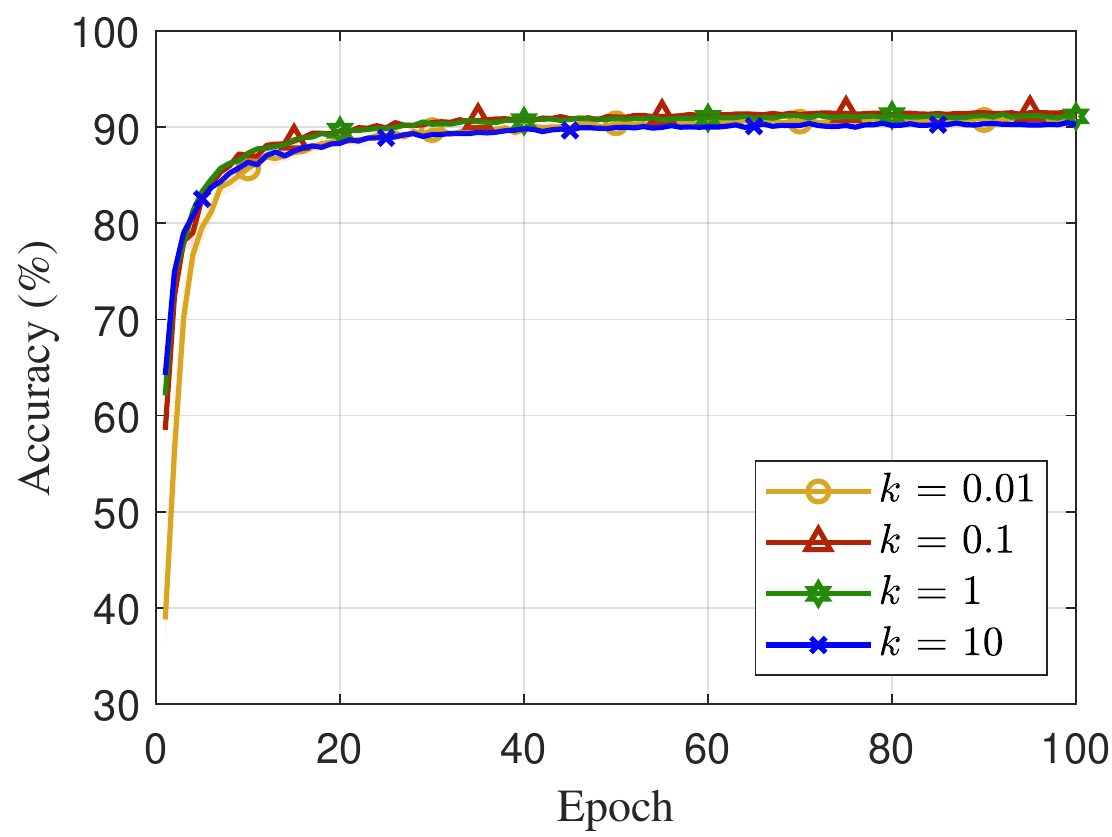}}
\subfigure[LSTMs on IMDB]{\includegraphics[width=0.32\textwidth]{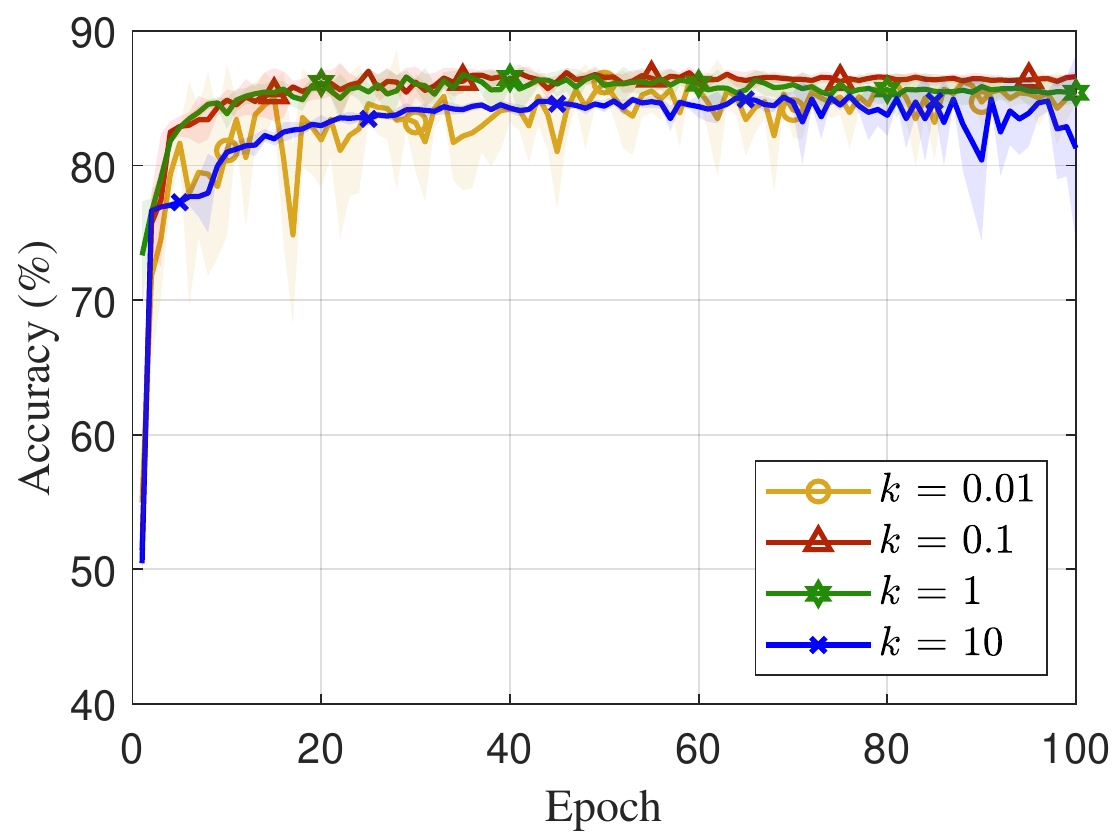}}
\caption{Influence investigation of the ratio factor $k$ on the performances of our algorithms used for MNIST, CIFAR-10 and IMDB with the MSE loss.}
\end{figure*}
\begin{figure*}[htbp]
\centering
\subfigure[FNNs on MNIST]{\includegraphics[width=0.32\textwidth]{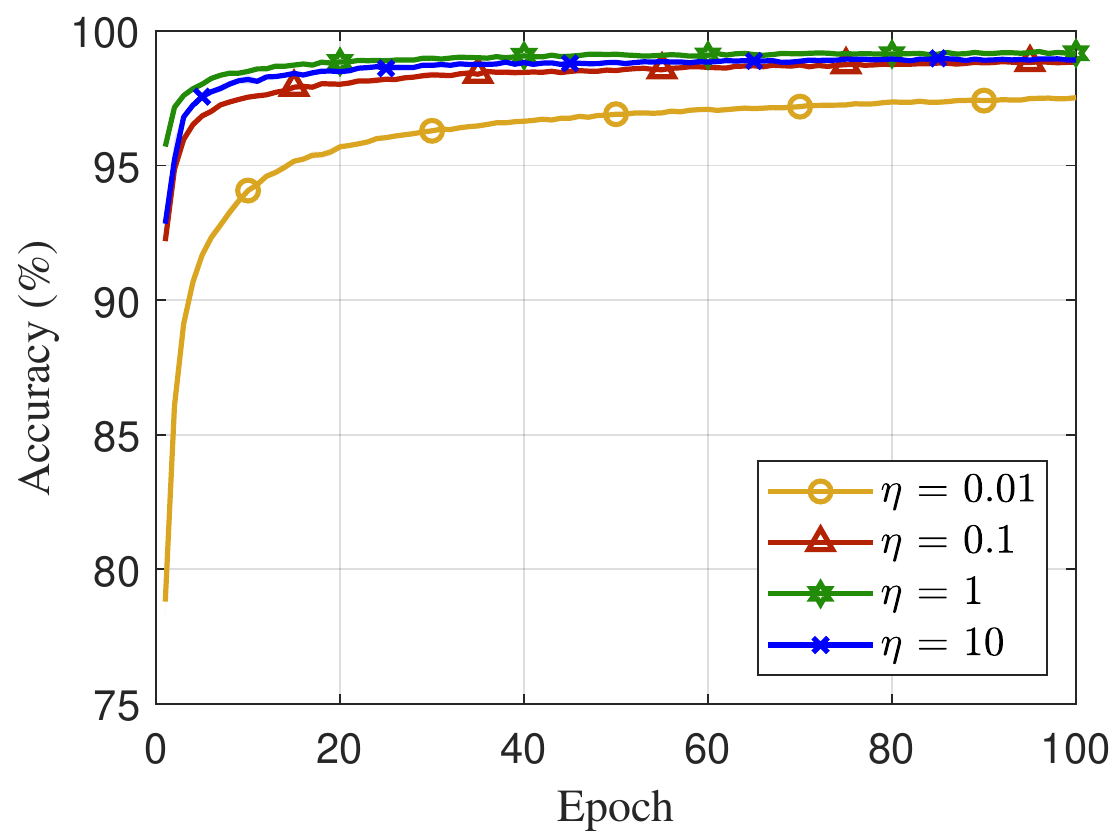}}
\subfigure[CNNs on CIFAR-10]{\includegraphics[width=0.32\textwidth]{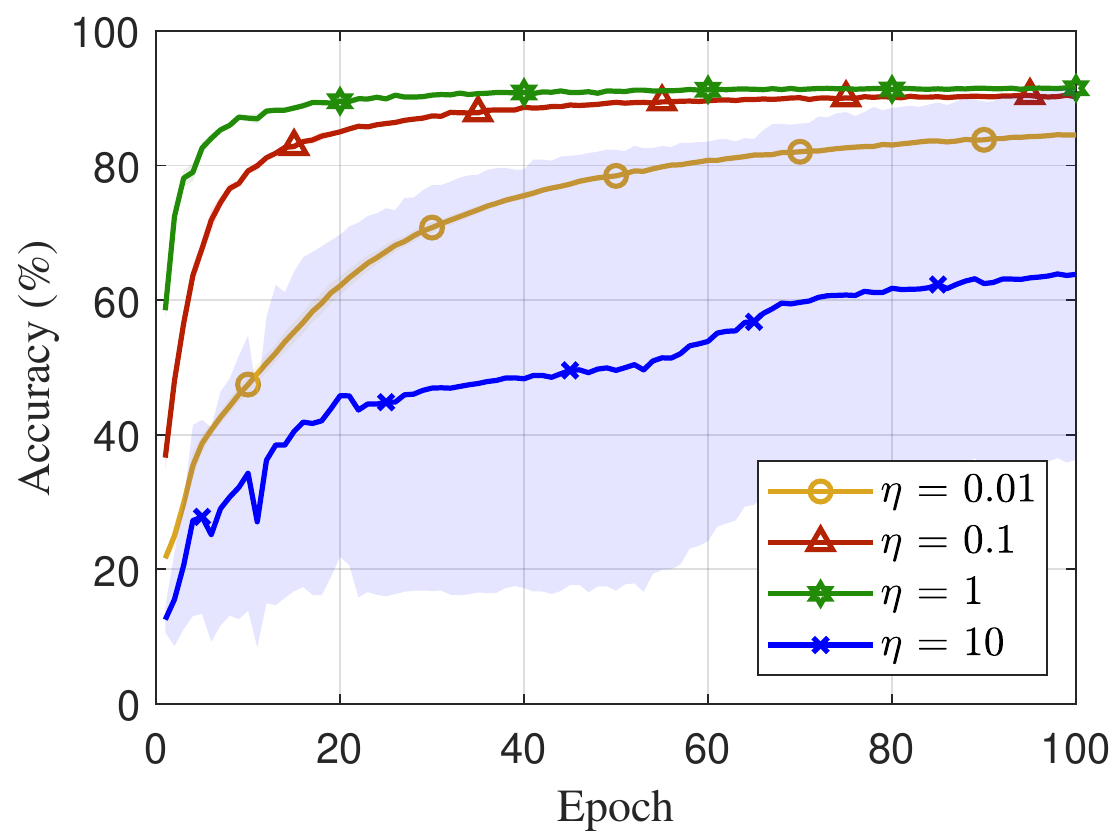}}
\subfigure[LSTMs on IMDB]{\includegraphics[width=0.32\textwidth]{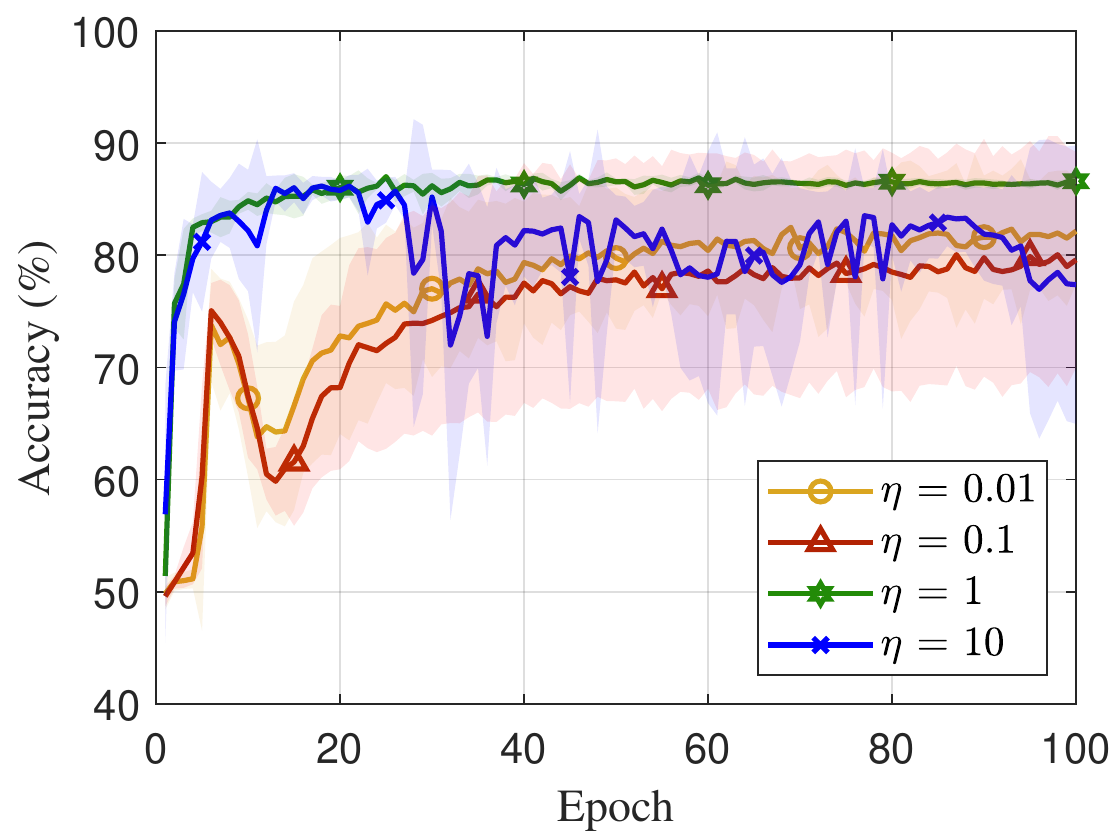}}
\caption{Influence investigation of the scaling factor $\eta$ on the performances of our algorithms used for MNIST, CIFAR-10 and IMDB with the MSE loss.}
\end{figure*}

The settings of all DNNs used here are summarized as follows: 1) Our FNNs consist of an input layer with 784 units, a ReLU FC hidden layer with 512 neurons and
a FC output layer with 10 neurons. 2) Our CNNs can be viewed as a mini VGG with five CONV layers and two FC layers. The five CONV layers are the same
as the first five CONV layers in VGG-16. The first FC layer is a ReLU hidden layer with 1024 neurons, and the last FC layer is the output layer with 10 neurons.
3) Our LSTMs consist of two stacked LSTM hidden layers with 512 units and an output layer with 2 neurons.  The activation functions
 $\sigma(\cdot)$ and $g(\cdot)$  are sigmoid($\cdot$) and tanh($\cdot$), respectively. In addition,
the output layer in all DNNs will use the identity activation function with the MSE loss defined by (57), if our algorithms
are used for training all layers. Otherwise, it will use the softmax function with the cross-entropy loss if our algorithms
are used only for hidden layers and Adam is used for the output layer.

Besides the above settings, other experimental settings are listed as follows:
1) All algorithms are tested five times on each dataset repeatedly, and
each time runs 100 epochs.
2) The minibatch size for all algorithms is fixed to 128.
3) For the performance evaluation of our algorithms,
all inverse autocorrelation matrices are initialized to identity matrices, and $\lambda$, $k$ and all gradient scaling factors are fixed
 to 1, 0.1 and 1, respectively. In addition, $\alpha$ will be
fixed to 0.5, and $\gamma$ will be fixed to $10^{-5}$ for FNNs and CNNs or $10^{-6}$ for LSTMs, if we use the improved methods presented in
Section \uppercase\expandafter{\romannumeral5}.\textit{B}. for our algorithms.
4) For the influence investigation of hyperparameters of our algorithms,
$k$ and all gradient scaling factors are selected from $\{0.01,0.1,1,10\}$ in turn, and other hyperparameters are the same as those in 3).
5) To avoid exploding gradients, we use the $L_2$ norm to clip gradients to 5.0 in FNNs and CNNs and to 1.0 in LSTMs.
6) All DNNs and algorithms are implemented by PyTorch and carried out on one 1080Ti GPU. The parameters of all DNNs and the
hyperparamters of Adam are initialized and set by default.

\subsection{Experimental Results}
The performance evaluation results of our algorithms against the Adam algorithm are summarized in Fig. 2 and  Fig. 3, where
legends ``HRLS", ``+M"  and ``+R" denote our algorithms together with Adam,
the momentum and the $L_1$ regularization, respectively. From Fig. 2 and  Fig. 3, we can draw the following conclusions:
1) Our algorithms have faster convergence speed and better convergence quality than Adam, especially for training CNNs.
2) By combing with the momentum and the $L_1$ regularization, our algorithms can obtain better performance.
3) Our algorithms can be used in combination with other first-order algorithms such as Adam, even if the loss
 function is not the MSE loss and the activation function of the output layer is not invertible.
 In addition, the running time of our algorithms for FNNs, CNNs and LSTMs is about 1.1, 1.4 and 1.1 times
 the running time of Adam for those, respectively.

The influence investigation results of hyperparameters $k$ and $\eta^l$ (or $\{(\eta^{l_w}, \eta^{l_v})\}_{l=1}^{L-1}$)
on the performances of our algorithms are presented in Fig. 4 and  Fig. 5.
From Fig. 4, the ratio factor $k$ should be set to 0.1. A too big or too small value for $k$ will
deteriorate the convergence quality of our algorithms. From Fig. 5, the gradient scaling factor $\eta$ should be set to 1.
A too big value for $\eta$ will cause gradient explosion, and a too small value for $\eta$ will
slow down the convergence speed. By comparing Fig. 4 with Fig. 5, we can easily find that our algorithms are more sensitive to
$\eta$ than $k$. From (53), (55), (84), (86), (94), (96), (62), (63) and (97),
a small $k$ will slow down the update of the inverse autocorrelation matrix and increase the learning rate
of network parameters indirectly, but a small $\eta$ will decrease the learning rate directly.
\section{Conclusion}
In this work, we revisited the RLS optimization technique for neural networks, and made it suitable for training
DNNs. We derived three RLS optimization algorithms for training FNNs, CNNs and RNNs (including LSTMs), respectively.
By using our proposed average-approximation and equivalent-gradient methods, we
converted our algorithms into a special SGD algorithm, which is simple and easy to use.
Complexity analysis shows that the time and space complexities of our algorithms are only several times those of
the conventional SGD algorithm. We further presented two improved methods for our algorithms.
Finally, we evaluated the convergence performances of our algorithms with FNNs on MNIST,
CNNs on CIFAR-10 and LSTMs on IMDB. Experiments results show that our algorithms have
faster convergence speed and better convergence quality than Adam. We also investigated the
influences of hyperparameters on the performances of our algorithms experimentally.
We hope that our work can help the RLS optimization for training neural networks flourish again,
although our work still has room for further improvement.


%

\section*{Acknowledgment}
This work is supported by the National Natural Science
Foundation of China (61762032, 11961018). The authors are
grateful to all anonymous reviewers for their constructive comments and valuable suggestions.
\ifCLASSOPTIONcaptionsoff
  \newpage
\fi



%

\bibliography{./RLSOPTDNNs}

\begin{thebibliography}{10}
\providecommand{\url}[1]{#1}
\csname url@samestyle\endcsname
\providecommand{\newblock}{\relax}
\providecommand{\bibinfo}[2]{#2}
\providecommand{\BIBentrySTDinterwordspacing}{\spaceskip=0pt\relax}
\providecommand{\BIBentryALTinterwordstretchfactor}{4}
\providecommand{\BIBentryALTinterwordspacing}{\spaceskip=\fontdimen2\font plus
\BIBentryALTinterwordstretchfactor\fontdimen3\font minus
  \fontdimen4\font\relax}
\providecommand{\BIBforeignlanguage}[2]{{%
\expandafter\ifx\csname l@#1\endcsname\relax
\typeout{** WARNING: IEEEtran.bst: No hyphenation pattern has been}%
\typeout{** loaded for the language `#1'. Using the pattern for}%
\typeout{** the default language instead.}%
\else
\language=\csname l@#1\endcsname
\fi
#2}}
\providecommand{\BIBdecl}{\relax}
\BIBdecl

\bibitem{Goodfellow2016}
I.~Goodfellow, Y.~Bengio, and A.~Courville, \emph{Deep Learning}.\hskip 1em
  plus 0.5em minus 0.4em\relax Cambridge, MA: MIT Press, 2016.

\bibitem{LeCunBH2015}
Y.~LeCun, Y.~Bengio, and G.~E. Hinton, ``Deep learning,'' \emph{Nature}, vol.
  521, no. 7553, pp. 436--444, 2015.

\bibitem{VoulodimosDDP2018}
A.~Voulodimos, N.~Doulamis, A.~Doulamis, and E.~Protopapadakis, ``Deep learning
  for computer vision: {A} brief review,'' \emph{Comp. Int. and Neurosc.}, vol.
  2018, pp. 7\,068\,349:1--7\,068\,349:13, 2018.

\bibitem{Malik2021}
M.~Malik, M.~K. Malik, K.~Mehmood, and I.~Makhdoom, ``Automatic speech
  recognition: A survey,'' \emph{Multim. Tools Appl.}, vol.~80, no.~6, pp.
  9411--9457, 2021.

\bibitem{Otter2021}
D.~W. Otter, J.~R. Medina, and J.~K. Kalita, ``A survey of the usages of deep
  learning for natural language processing,'' \emph{{IEEE} Trans. Neural
  Networks Learn. Syst.}, vol.~32, no.~2, pp. 604--624, 2021.

\bibitem{mu2018}
R.~Mu, ``A survey of recommender systems based on deep learning,'' \emph{{IEEE}
  Access}, vol.~6, pp. 69\,009--69\,022, 2018.

\bibitem{Esteva2019}
A.~Esteva, A.~Robicquet, B.~Ramsundar, V.~Kuleshov, M.~DePristo, K.~Chou,
  C.~Cui, G.~Corrado, S.~Thrun, and J.~Dean, ``A guide to deep learning in
  healthcare,'' \emph{Nature Medicine}, vol.~25, no.~1, pp. 24--29, Jan. 2019.

\bibitem{Bottou1991}
L.~Bottou, ``Stochastic gradient learning in neural networks,'' in \emph{Proc.
  of 1991 Neuro-N{\^i}mes}, Nimes, France, 1991.

\bibitem{Sutskever2013}
I.~Sutskever, J.~Martens, G.~E. Dahl, and G.~E. Hinton, ``On the importance of
  initialization and momentum in deep learning,'' in \emph{Proc. of 30th Int.
  Conf. Mach. Learn.}, Atlanta, GA, USA, Jun. 2013, pp. 1139--1147.

\bibitem{Duchi2011}
J.~Duchi, E.~Hazan, and Y.~Singer, ``Adaptive subgradient methods for online
  learning and stochastic optimization,'' \emph{J. Mach. Learn. Res.}, vol.~12,
  pp. 2121--2159, 2011.

\bibitem{Tieleman2012}
T.~Tieleman and G.~Hinton, ``Lecture 6e rmsprop: Divide the gradient by a
  running average of its recent magnitude,'' \emph{COURSERA: Neural Networks
  for Mach. Learn.}, vol.~4, pp. 26--30, 2012.

\bibitem{Kingma2015}
D.~P. Kingma and J.~L. Ba, ``Adam: {A} method for stochastic optimization,'' in
  \emph{Proc. of 3rd Int. Conf. Learn. Representations}, San Diego, CA, USA,
  May 2015.

\bibitem{Wilson2017}
A.~C. Wilson, R.~Roelofs, M.~Stern, N.~Srebro, and B.~Recht, ``The marginal
  value of adaptive gradient methods in machine learning,'' in \emph{Proc. of
  31st Neural Inf. Process. Syst.}, Long Beach, CA, USA, Dec. 2017, pp.
  4148--4158.

\bibitem{martens2016}
J.~Martens, ``Second-order optimization for neural networks,'' Ph.D.
  dissertation, University of Toronto, Toronto, Canada, 2016.

\bibitem{Battiti1992}
R.~Battiti, ``First- and second-order methods for learning: Between steepest
  descent and newton's method,'' \emph{Neural Computation}, vol.~4, no.~2, pp.
  141--166, 1992.

\bibitem{Chen2018}
C.~P. H and H.~Cho-jui, ``A comparison of second-order methods for deep
  convolutional neural networks,'' in \emph{Proc. of 6th Int. Conf. Learn.
  Representations}, Vancouver, BC, Canada, May 2018.

\bibitem{Martens2010}
J.~Martens, ``Deep learning via hessian-free optimization,'' in \emph{Proc. of
  27th Int. Conf. Mach. Learn.}, Haifa, Israel, 2010, pp. 735--742.

\bibitem{Martens2011}
J.~Martens and I.~Sutskever, ``Learning recurrent neural networks with
  hessian-free optimization,'' in \emph{Proc. of 28th Int. Conf. Mach. Learn.},
  Bellevue, Washington, USA, 2011, pp. 1033--1040.

\bibitem{Martens2015}
J.~Martens and R.~B. Grosse, ``Optimizing neural networks with
  kronecker-factored approximate curvature,'' in \emph{Proc. of 32nd Int. Conf.
  Mach. Learn.}, Lille, France, 2015, pp. 2408--2417.

\bibitem{Botev2017}
A.~Botev, H.~Ritter, and D.~Barber, ``Practical gauss-newton optimisation for
  deep learning,'' in \emph{Proc. of 34th Int. Conf. Mach. Learn.}, Sydney,
  NSW, Australia, 2017, pp. 557--565.

\bibitem{Wang2017}
X.~Wang, S.~Ma, D.~Goldfarb, and W.~Liu, ``Stochastic quasi-newton methods for
  nonconvex stochastic optimization,'' \emph{SIAM J. on Optim.}, vol.~27,
  no.~2, pp. 927--956, 2017.

\bibitem{Goldfarb2020}
D.~Goldfarb, Y.~Ren, and A.~Bahamou, ``Practical quasi-newton methods for
  training deep neural networks,'' in \emph{Proc. of 34th Neural Inf. Process.
  Syst.}, Vancouver, BC, Canada, Dec. 2020.

\bibitem{Xu2020}
P.~Xu, F.~Roosta, and M.~W. Mahoney, ``Newton-type methods for non-convex
  optimization under inexact hessian information,'' \emph{Math. Program.}, vol.
  184, no.~1, pp. 35--70, 2020.

\bibitem{Yam1995}
J.~Y.~F. Yam and T.~W.~S. Chow, ``Accelerated training algorithm for
  feedforward neural networks based on least squares method,'' \emph{Neural
  Process. Lett.}, vol.~2, no.~4, pp. 20--25, 1995.

\bibitem{Ergezinger1995}
S.~Ergezinger and E.~Thomsen, ``An accelerated learning algorithm for
  multilayer perceptrons: Optimization layer by layer,'' \emph{{IEEE} Trans.
  Neural Networks}, vol.~6, no.~1, pp. 31--42, 1995.

\bibitem{Yam1997}
J.~Y.~F. Yam and T.~W.~S. Chow, ``Extended least squares based algorithm for
  training feedforward networks,'' \emph{{IEEE} Trans. Neural Networks},
  vol.~8, no.~3, pp. 806--810, 1997.

\bibitem{Biegler-Konig1993}
F.~Biegler{-}K{\"{o}}nig and F.~B{\"{a}}rmann, ``A learning algorithm for
  multilayered neural networks based on linear least squares problems,''
  \emph{Neural Networks}, vol.~6, no.~1, pp. 127--131, 1993.

\bibitem{Fontenla-Romero2003}
O.~Fontenla{-}Romero, D.~Erdogmus, J.~C. Pr{\'{\i}}ncipe, A.~Alonso{-}Betanzos,
  and E.~F. Castillo, ``Linear least-squares based methods for neural networks
  learning,'' in \emph{Proc. of 2003 Artif. Neural Networks and Neural Inf.
  Process.}, vol. 2714, Istanbul, Turkey, 2003, pp. 84--91.

\bibitem{Azimi-Sadjadi1990}
M.~R. Azimi{-}Sadjadi, S.~Citrin, and S.~Sheedvash, ``Supervised learning
  process of multi-layer perceptron neural networks using fast least squares,''
  in \emph{Proc. of 1990 Int. Conf. Acoust., Speech, and Signal Process.},
  Albuquerque, New Mexico, USA, 1990, pp. 1381--1384.

\bibitem{Azimi-Sadjadi1992}
M.~R. Azimi{-}Sadjadi and R.~Liou, ``Fast learning process of multilayer neural
  networks using recursive least squares method,'' \emph{{IEEE} Trans. Signal
  Process.}, vol.~40, no.~2, pp. 446--450, 1992.

\bibitem{Cho2001}
S.~Cho, T.~W.~S. Chow, and Y.~Fang, ``Training recurrent neural networks using
  optimization layer-by- layer recursive least squares algorithm for vibration
  signals system identification and fault diagnostic analysis,'' \emph{J. of
  Intell. Syst.}, vol.~11, no.~2, pp. 125--154, 2001.

\bibitem{Al-Batah2010}
M.~S. Al{-}Batah, N.~A.~M. Isa, K.~Z. Zamli, and K.~A. Azizli, ``Modified
  recursive least squares algorithm to train the hybrid multilayered perceptron
  {(HMLP)} network,'' \emph{Appl. Soft Comput.}, vol.~10, no.~1, pp. 236--244,
  2010.

\bibitem{Stan2000}
O.~Stan and E.~W. Kamen, ``A local linearized least squares algorithm for
  training feedforward neural networks,'' \emph{{IEEE} Trans. Neural Netw.
  Learn Syst.}, vol.~11, no.~2, pp. 487--495, 2000.

\bibitem{Bilski1998}
J.~Bilski and L.~Rutkowski, ``A fast training algorithm for neural networks,''
  \emph{{IEEE} Trans. on Circuits and Syst.II: Analog and Digital Signal
  Process.}, vol.~45, no.~6, pp. 749--753, 1998.

\bibitem{Huang2006}
G.~Huang, L.~Chen, and C.~K. Siew, ``Universal approximation using incremental
  constructive feedforward networks with random hidden nodes,'' \emph{{IEEE}
  Trans. on Neural Networks}, vol.~17, no.~4, pp. 879--892, 2006.

\bibitem{Huang2012}
G.~Huang, H.~Zhou, X.~Ding, and R.~Zhang, ``Extreme learning machine for
  regression and multiclass classification,'' \emph{{IEEE} Trans. Syst. Man
  Cybern. Part {B}}, vol.~42, no.~2, pp. 513--529, 2012.

\bibitem{Zhou2015}
H.~Zhou, G.~Huang, Z.~Lin, H.~Wang, and Y.~C. Soh, ``Stacked extreme learning
  machines,'' \emph{{IEEE} Trans. Cybern.}, vol.~45, no.~9, pp. 2013--2025,
  2015.

\bibitem{Pang2016}
S.~Pang and X.~Yang, ``Deep convolutional extreme learning machine and its
  application in handwritten digit classification,'' \emph{Comput. Intell.
  Neurosci.}, vol. 2016, pp. 3\,049\,632:1--3\,049\,632:10, 2016.

\bibitem{Park2017}
J.~Park and J.~Kim, ``Online recurrent extreme learning machine and its
  application to time-series prediction,'' in \emph{Proc. of 2017 Int. Joint
  Conf. on Neural Networks}, Anchorage, AK, USA, May 2017, pp. 1983--1990.

\bibitem{Jaeger2002}
H.~Jaeger, ``A tutorial on training recurrent neural networks, covering bppt,
  rtrl, ekf and the `echo state network' approach,'' German National Research
  Center for Information Technology, Sankt Augustin, Germany, GMD Report 159,
  2002.

\bibitem{Hochreiter1997}
S.~Hochreiter and J.~Schmidhuber, ``Long short-term memory,'' \emph{Neural
  Comput.}, vol.~9, no.~8, pp. 1735--1780, 1997.

\bibitem{Petersen2012}
K.~B. Petersen and M.~S. Pedersen, \emph{The Matrix Cookbook}.\hskip 1em plus
  0.5em minus 0.4em\relax Technical University of Denmark, 2012.

\bibitem{Claser2016}
R.~Claser and V.~H.~N. andYuriy V.~Zakharov, ``A low-complexity {RLS-DCD}
  algorithm for volterra system identification,'' in \emph{Proc. of 24th
  European Signal Processing Conference}, Budapest, Hungary, Aug. 2016, pp.
  6--10.

\bibitem{Sadigh2020}
A.~N. Sadigh, A.~H. Taherinia, and H.~S. Yazdi, ``Analysis of robust recursive
  least squares: Convergence and tracking,'' \emph{Signal Process.}, vol. 171,
  p. 107482, 2020.

\bibitem{Eksioglu2010}
E.~M. Eksioglu, ``{RLS} adaptive filtering with sparsity regularization,'' in
  \emph{Proc. of 10th Int. Conf. on Inf. Sciences, Signal Process. and their
  Appl.}, Kuala Lumpur, Malaysia, May 2010, pp. 550--553.

\bibitem{Albu2012}
F.~Albu, ``Improved variable forgetting factor recursive least square
  algorithm,'' in \emph{Proc. of 12th Int. Conf. on Control Automat. Robot. \&
  Vision}, Guangzhou, China, Dec. 2012, pp. 1789--1793.

\bibitem{Eksioglu2011}
E.~M. Eksioglu and A.~K. Tanc, ``{RLS} algorithm with convex regularization,''
  \emph{{IEEE} Signal Process. Lett.}, vol.~18, no.~8, pp. 470--473, 2011.

\bibitem{Polyak1964}
B.~T. Polyak, ``Some methods of speeding up the convergence of iteration
  methods,'' \emph{USSR Computational Mathematics and Mathematical Physics},
  vol.~4, no.~5, pp. 1--17, 1964.

\end{thebibliography}
%








\end{document}